\documentclass[10pt,journal,compsoc]{IEEEtran}
\ifCLASSOPTIONcompsoc
\else
\fi
\usepackage[utf8]{inputenc} 
\usepackage[T1]{fontenc}    
\usepackage{hyperref}       
\usepackage{url}            
\usepackage{booktabs}       
\usepackage{amsfonts}       
\usepackage{nicefrac}       
\usepackage{microtype}      
\usepackage{xcolor}         
\usepackage{graphicx}
\usepackage{subfigure}
\usepackage{amsmath,amssymb} 
\usepackage{caption} 
\usepackage{hyperref}
\usepackage{wrapfig}
\usepackage[american]{babel}
\usepackage{paralist,algorithm}
\usepackage[noend]{algpseudocode}

\newcommand{\x}{{\bf x}}
\newcommand{\eg}{{\em e.g.}}
\newcommand{\ie}{{\em i.e.}}

\newcommand{\p}{{\bf p}}

\newcommand{\y}{{\bf y}}

\newcommand{\name}{{\sc LeadNet}}
\newtheorem{remark}{Remark}

\ifCLASSINFOpdf
\else
\fi

\begin{document}
%
\title{Contextualizing Meta-Learning \\via Learning to Decompose}
%
%
%
%

\author{Han-Jia Ye,
	    Da-Wei Zhou,
	    Lanqing Hong,
	    Zhenguo Li,
	    Xiu-Shen Wei,
        De-Chuan Zhan
\IEEEcompsocitemizethanks{
\IEEEcompsocthanksitem H.-J. Ye, D.-W. Zhou, and D.-C. Zhan are with State Key Laboratory for Novel Software Technology, Nanjing University,  Nanjing, 210023, China. E-mail: \{yehj, zhoudw, zhandc\}@lamda.nju.edu.cn.
\IEEEcompsocthanksitem L. Hong and Z. Li are with Huawei Noah’s Ark Lab. E-mail: \{honglanqing, li.zhenguo\}@huawei.com.
\IEEEcompsocthanksitem X.-S. Wei is with the School of Computer Science and Engineering, Southeast University, Nanjing, 210096, China, and Key Laboratory of New Generation Artificial Intelligence Technology and Its Interdisciplinary Applications (Southeast University), Ministry of Education, Nanjing, 210096, China. E-mail: weixs.gm@gmail.com.
\IEEEcompsocthanksitem (Corresponding author: De-Chuan Zhan.)}

}

%
%

\markboth{Journal of \LaTeX\ Class Files,~Vol.~Xx, No.~X, Xxxx~20Xx}%
{Ye \MakeLowercase{\textit{et al.}}: Contextualizing Multiple Tasks via Learning to Decompose}
%



\IEEEtitleabstractindextext{%
\begin{abstract}
Meta-learning has emerged as an efficient approach for constructing target models based on support sets. For example, the meta-learned embeddings enable the construction of target nearest-neighbor classifiers for specific tasks by pulling instances closer to their same-class neighbors.
However, a single instance can be annotated from various latent attributes, making visually similar instances inside or across support sets have different labels and diverse relationships with others. 
Consequently, a uniform meta-learned strategy for inferring the target model from the support set fails to capture the instance-wise ambiguous similarity.
To this end, we propose Learning to Decompose Network~({\name}) to {\em contextualize} the meta-learned ``support-to-target'' strategy, leveraging the context of instances with one or mixed latent attributes in a support set.
In particular, the comparison relationship between instances is decomposed w.r.t. multiple embedding spaces. {\name} learns to automatically select the strategy associated with the right attribute via incorporating {\em the change of comparison across contexts} with polysemous embeddings.
We demonstrate the superiority of {\name} in various applications, including exploring multiple views of confusing data, out-of-distribution recognition, and few-shot image classification.

\end{abstract}

}

\maketitle


%
\IEEEpeerreviewmaketitle

\IEEEraisesectionheading{\section{Introduction}}
\label{sec:intro}
Meta-learning has proven to be an effective approach for encoding a generalizable ``learning strategy from a support set to its classifier''~\cite{vilalta2002perspective,Chao2019Meta,hospedales2021meta}, which has been widely applied in various fields, including few-shot learning~\cite{VinyalsBLKW16Matching,FinnAL17Model,coskun2021domain,baik2021learning,sun2020meta}, object detection~\cite{kj2021incremental,xiao2019online,xiao2022few}, and recommendation systems~\cite{vartak2017meta,luo2020metaselector,deng2022new}. 
Given a large number of labeled classes (a.k.a. {\em base} classes), a meta-learner samples episodes of (pseudo) tasks from them to mimic the target deployment scenario, and optimizes the ``learning strategy'' to construct an effective classifier based on the support set.
Typically, a support set is associated with a query set within a (pseudo) task, sharing the same set of classes, to evaluate the model output by the meta-learner. 
Subsequently, the meta-model is applied to tasks with non-overlapping {\em novel} classes, enabling the classifier constructed by the meta-model to recognize new instances of these novel classes. 
One common implementation of the meta-model is through an embedding function (the feature extractor), where the ``learning strategy'' becomes an embedding-based classifier~\cite{VinyalsBLKW16Matching,SnellSZ17Prototypical,Lee2019Meta}.
Additionally, the meta-model can also be optimizers~\cite{FinnAL17Model,Nichol2018On,ye2020few} and image generators~\cite{Wang2018Low}.

The semantic meanings of instances play a crucial role in both sampling (pseudo) tasks in meta-learning and constructing a generalizable meta-model. 
However, instances often exhibit rich semantics, and their labels can vary significantly due to different annotators' preferences~\cite{changpinyo2013similarity,tan2019learning,Vuorio2019Multimodal,Su2020Task}.
For example, the label of an object like ``red boots'' could be annotated based on different latent attributes. One annotator may choose to label it as ``boots'', emphasizing the object's ``category'' attribute, while another annotator may prioritize the ``color'' aspect and label it as ``red''.
In the presence of annotations caused by multiple latent attributes, a meta-learner may mistakenly treat labels from different attribute levels equally. Consequently, the inconsistent labels assigned to visually similar instances can potentially mislead the meta-model's update and affect its performance.

\begin{figure*}[t]
	\centering
	\begin{tabular}{c|c}
		\includegraphics[height=0.24\textheight]{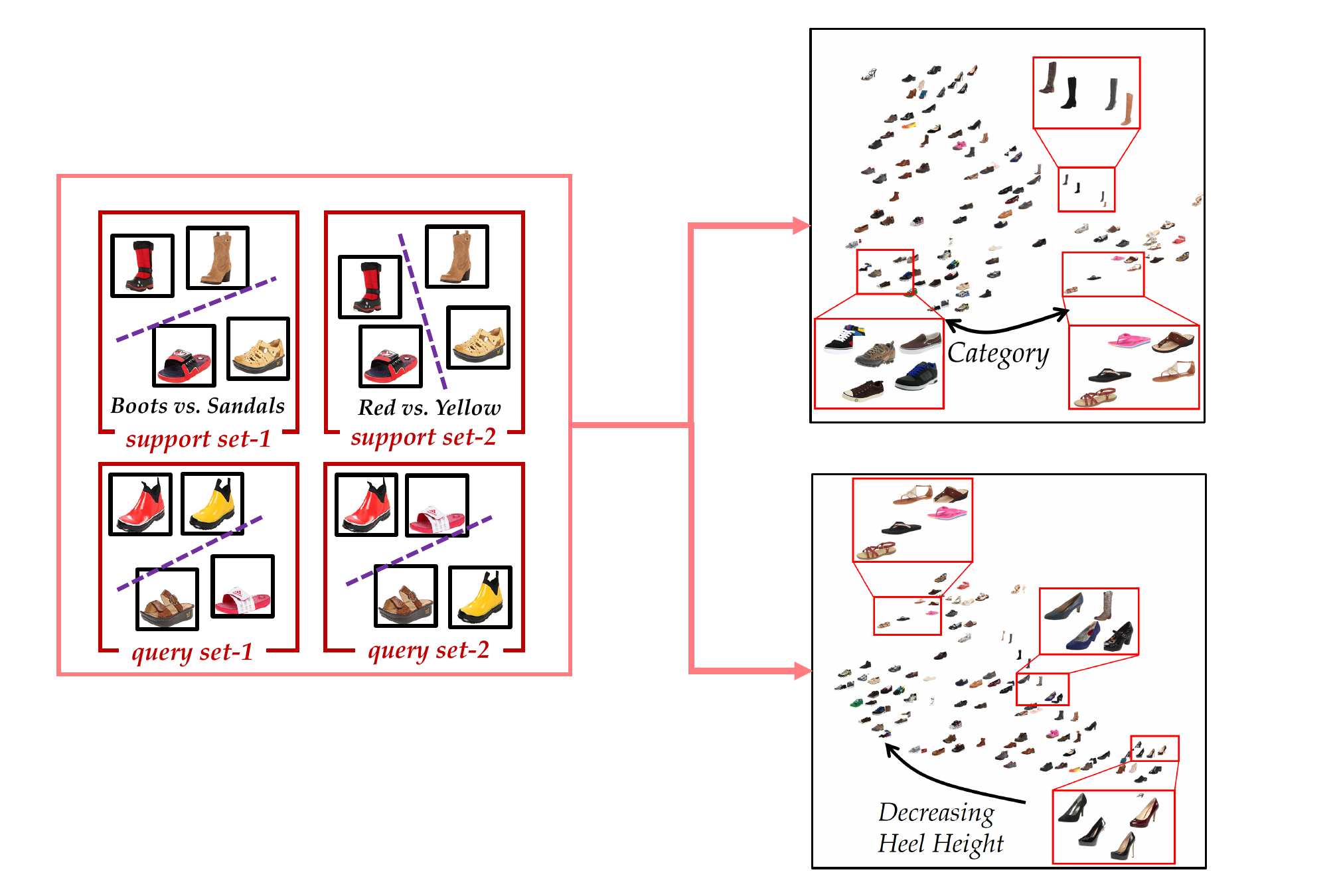} & \includegraphics[height=0.24\textheight]{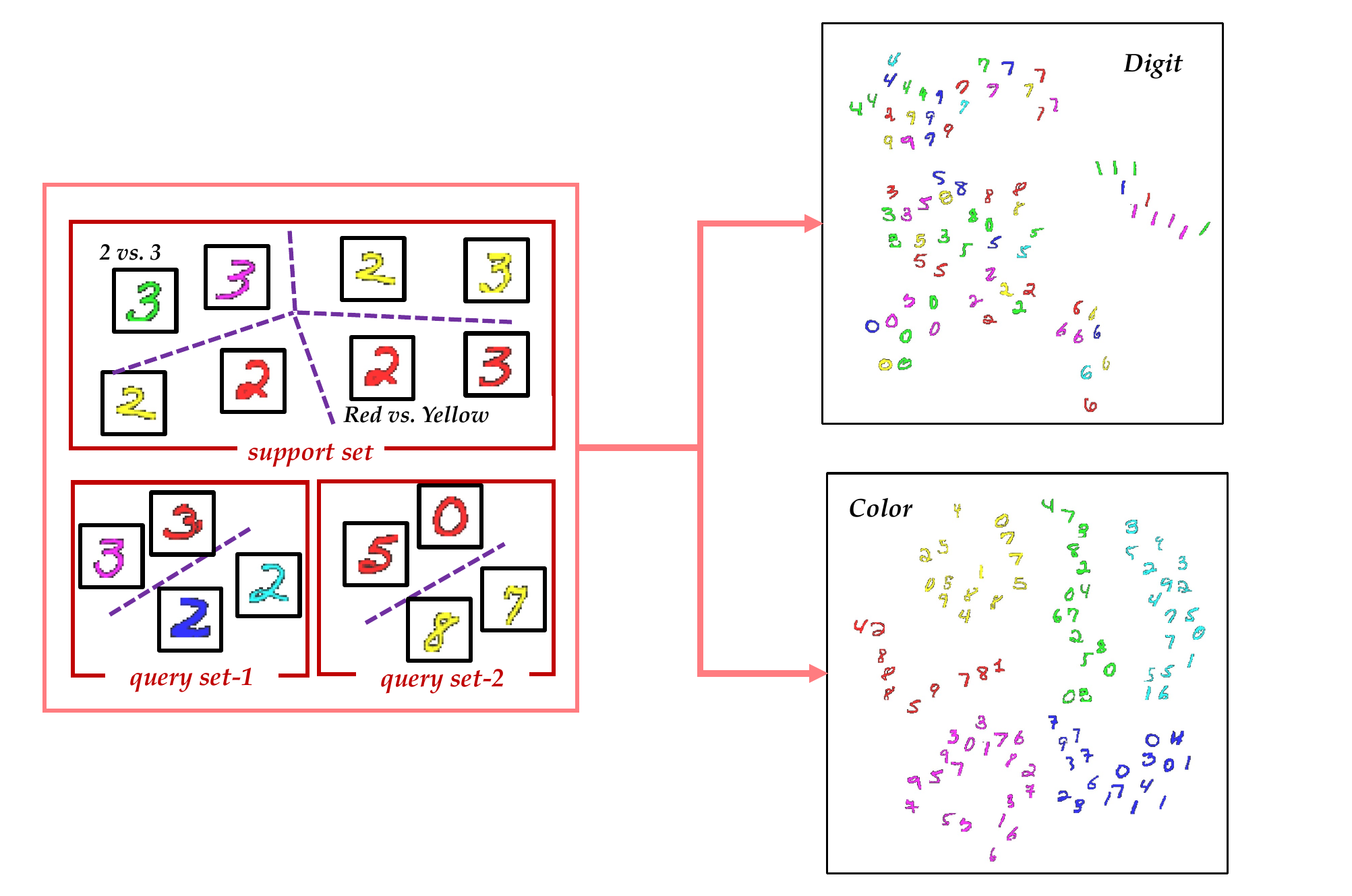} \\
		\textbf{(a)} Single Attribute per Task (SAT) & \textbf{(b)} Mixed Attributes per Task (MAT)
	\end{tabular}
	\caption{
		An illustration of the Single Attribute per Task (SAT) and the Mixed Attributes per Task (MAT) scenarios. We also show the t-SNE~\cite{van2008visualizing} of the decomposed multiple embedding spaces by {\name}. Detailed configurations and discussions are in Section~\ref{sec:experiment}.
(a) SAT. We take UT-Zappos-50k dataset as an example. Instances in a support set are annotated from a certain aspect (with a specific latent attribute), such as ``category'' or ``heel height'', and the same instance could be labeled differently in two tasks. A contextualized model learned from the support set is asked to discern instances in the corresponding query set sharing the same attribute.
(b) MAT. We show the results of {\name} on Color Digits, where the 4-class instances in the support set are labeled based on two latent attributes (``color'' and ``digit''). A contextualized model should capture both attributes and handle two query sets with instances labeled by corresponding attributes.
	}
	\label{fig:semantics}
\end{figure*}

There are two challenging cases when instances in meta-learning are labeled with multiple {\em unknown latent} attributes. 
First, all instances in a task are annotated based on a specific attribute, but the labels of the same instance may differ across tasks.\footnote{In the remainder of the paper, we use ``task'' and ``(pseudo) task'' in an exchangeable manner, denoting a pair of support and query sets.} For example, in Fig.~\ref{fig:semantics} (a), two pairs of tasks are related to  ``category'' and ``color'' attributes, respectively. 
Second, instances are directly annotated with mixed attributes. As illustrated in Fig.~\ref{fig:semantics} (b), instances may be labeled with ``digit'' labels like ``2'' and ``3'' or ``color'' labels such as ``red'' and ``yellow''.
These scenarios introduce ambiguity and inconsistency in labeling, making it difficult to devise a uniform learning strategy for the meta-model. Hence, training a monosemous meta-model without considering the latent attributes may limit its generalization ability.
While some methods adapt the model based on the given support set to handle variety~\cite{Oreshkin2018TADAM,Yao2019Hierarchically,Zintgraf2019Fast,Vuorio2019Multimodal}, they fail to address the mixed attribute case and overlook the relationships between instances. 
In addition, multi-label meta-learning methods~\cite{simon2022meta,wu2019learning} require access to ground truth attributes, which is not feasible in both of the aforementioned scenarios.

To deal with the polysemous objects in meta-learning, we propose LEArning to Decompose Network~(\name) to contextualize the meta-model. 
We assume the existence of multiple ``support-to-target'' that are associated with the possible attributes, and our objective is to enable the meta-model to identify the specific latent attribute present in the instances and utilize the corresponding strategy accordingly.
{\name} is {\em general}  for both cases when instances are labeled by one single or mixed confusing attributes within a task.
By decomposing the meta-learned strategies, {\name} facilitates the discovery of latent attributes, enhancing the model's adaptability and making it {\em widely applicable} to a diverse range of ``downstream'' tasks with rich semantics.

The main challenge in contextualized meta-learning lies in effectively incorporating semantics during the meta-training phase. 
We propose to analyze the semantics conditioned on a {\em context} --- a set of instances within or across the support set. 
Instead of treating instances separately, in {\name}, we link the comparison relationships of instances across tasks to help infer the variety of latent attributes.
If one instance exhibits diverse comparison relationships with others across tasks, \eg, ``red boots'' is similar to ``yellow boots'' in a task but has the same label as ``red sandals'' in another task, it indicates that two attributes should be considered to explain the semantic variability. 
Thus, {\name} meta-learns multiple embedding spaces corresponding to the latent attributes. This helps construct multiple meta-learned strategies from the support set to the target model and facilitates decomposing the comparison relationship between instances.

Specifically, {\name} takes into account multiple attributes, each associated with a meta-learned strategy, when measuring the similarity between instances during the discerning process of the constructed target model.
{\name} then leverages the comparisons between instances within a support set as the context to determine the appropriate strategy, which tailor the meta-model based on the underlying context information.

{\name} is {\em flexible} for a number of applications.
First, it is able to decompose a curve with mixed function families and recover meaningful latent components with interpretability; 
the same ability also facilitates similarity learning, where the relationship between objects varies from task to task; 
furthermore, {\name} is able to generalizes its discerning ability to {\em out-of-distribution} examples and tasks; 
finally, {\name} achieves promising results on few-shot classification benchmarks.

Our contributions could be summarized as follows:
\begin{itemize}
\item When the latent attribute changes from task to task with limited instances, we claim the need to contextualize a meta-model to deal with the ambiguously labeled instances inside or across episodes of (pseudo) tasks;
\item We propose a general approach {\name} to infer the latent attribute of a set of instances via leveraging the change of their comparison relationships. {\name} is applicable when instances in a support set are labeled based on one or more latent attributes;
\item Benefiting from the ability to discover embedding spaces corresponding to latent attributes, {\name} facilitates applications like confusing multi-task learning, out-of-distribution recognition, and few-shot classification.
\end{itemize}

We present {\name} after the background, including implementations of the variants for two confusing scenarios. Finally are related work, experiments, and conclusions.
\section{Notations and Background}
\label{sec:problem}
We consider learning a classifier from a support set $\mathcal{S} = \{(\x_i, \y_i)\}_{i=1}^{NK}$, \ie, a set of examples with $N$ classes and $K$ instances in each of those classes. An instance $\x_i$ is coupled with a label $y_i\in[N]=\{1,\ldots,N\}$. $\y_i\in\{0,1\}^N$ is the one-hot coding of $y_i$. 
The goal is to capture the characteristic of $\mathcal{S}$, where the learned classifier predicts the label for those instances in $\mathcal{S}$ correctly. Besides, the classifier should also generalize its discerning ability to new instances in a query set $\mathcal{Q}$ sampled from the same distribution as $\mathcal{S}$. 
One way to predict $y_i$ given $\x_i$ is to estimate the posterior probability $\Pr(y_i=n \mid \x_i, \Theta)$ of the $n\in[N]$-th class based on $\x_i$ and the classifier parameter $\Theta$. We expect the ground truth class $y_i$ would have the largest probability with $\Theta$.

Meta-learning has become a useful tool to improve the quality of the classifier $\Theta$, especially when the size of $\mathcal{S}$ is small.
The key insight is to share classifier construction experience among a variety of tasks. 
We first introduce vanilla meta-learning and then define two scenarios where instances in meta-learning possess various latent attributes.

\subsection{Vanilla Meta-Learning}
\noindent\textbf{Meta-learning for classification.}
Given a {\em base} class set $\mathcal{B}$ (a.k.a. the meta-training set) collected in advance, meta-learning samples episodes of (pseudo) tasks, \ie, ($\mathcal{S}$, $\mathcal{Q}$) pairs.
In detail, $N$ classes are randomly chosen from the base class set $\mathcal{B}$. Then, corresponding instances are sampled and organized into support set $\mathcal{S}$ and query set $\mathcal{Q}$. 

A generalizable mapping $f$ from a support set to its corresponding classifier is extracted by the meta-learner over these episodes of tasks.
$f$ encodes the ``{\em learning strategy from a support set to its target classifier}'' over both those sampled (pseudo) tasks and even tasks with {\em novel} classes so that given a support set $\mathcal{S}$ we can construct a classifier more effectively with the generated parameter $\Theta = f(\mathcal{S})$.
Since a classifier derived from a good ``learning strategy'' predicts all query instances $\x\in\mathcal{Q}$ well in a certain task, we measure the quality of $f$ by the joint likelihood of query instances over all tasks:
\begin{equation}
\max_f \prod_{(\mathcal{S}, \mathcal{Q})\sim\mathcal{B}}\; \prod_{(\x, y)\in\mathcal{Q}}  \Pr(\hat{y} = y \mid \x, \; \mathcal{S}, \;\Theta=f(\mathcal{S}))\;\label{eq:meta-learning}\;.
\end{equation}
The prediction $\hat{y}$ depends on $\mathcal{S}$ and $\Theta$ jointly. All tasks share the same meta-model $f$.
Since $f$ fits different configurations of tasks, it is expected to generalize its discerning ability to new tasks (a.k.a. meta-test set), \eg, tasks with {\em novel} classes. Eq.~\ref{eq:meta-learning} becomes a popular way to learn a {\em few-shot} model where $K$ is small in $\mathcal{S}$~\cite{VinyalsBLKW16Matching,FinnAL17Model}.
We can reformulate Eq.~\ref{eq:meta-learning} in the form of minimizing the negative log-likelihood of predicted class labels for query instances:
\begin{align}
&\min_f \;\sum_{(\mathcal{S}, \mathcal{Q})\sim\mathcal{B}}\; \sum_{(\x, y)\in\mathcal{Q}}  -\log \Pr(\hat{y} = y \mid \x, \; \mathcal{S}, \;\Theta=f(\mathcal{S}))\notag\\
&\;=\;\sum_{(\mathcal{S}, \mathcal{Q})\sim\mathcal{B}}\; \sum_{(\x, y)\in\mathcal{Q}}  \ell(y, \;\hat{y})\;.\label{eq:meta-learning2}
\end{align}
Here, the discrepancy between the predicted label $\hat{y}$ and the true label $y$ could be measured by the loss $\ell(\cdot, \cdot)$, \eg, the cross-entropy.
The mapping $f$ of the meta-model could be implemented as the embedding function $\phi$, which extracts features and maps an object $\x$ from the input to a $d$-dimensional space. Thus, $\hat{y}$ is predicted based on the labels of support set neighbors measured by $\phi$. 
Denote $\mathbf{Dist}(\cdot, \cdot)$ as the distance between embeddings, then
\begin{align}
&\Pr(\hat{y} = y \mid \x, \mathcal{S},\phi) \notag\\
&\quad= \sum_{{\x_i\in\mathcal{S} \land y_i=y}} \frac{\exp\left(-\mathbf{Dist}(\phi(\x),\; \phi(\x_i))\right)}{\sum_{\x_j\in\mathcal{S}} \exp\left(-\mathbf{Dist}(\phi(\x),\;\phi(\x_j))\right)}\label{eq:embedding_prediction}\;.
\end{align}
Eq.~\ref{eq:embedding_prediction} depends on the closeness between $\phi(\x)$ and embeddings of class $y$ in $\mathcal{S}$. The smaller the distance, the larger the probability. By integrating Eq.~\ref{eq:embedding_prediction} with Eq.~\ref{eq:meta-learning}, the embedding $\phi$ pulls the same-class instances in the corresponding support set close and pushes other instances away. Owing to the fact that the average likelihood among multiple configurations of tasks is maximized, the learned embedding $\phi$ fits not only the base class tasks but also the tasks with novel classes. 

We implement $\mathbf{Dist}(\cdot, \cdot)$ in Eq.~\ref{eq:embedding_prediction} with negative cosine similarity~\cite{VinyalsBLKW16Matching}, and use the center of same class instances in $\mathcal{S}$ as a more powerful class representation when $K>1$~\cite{SnellSZ17Prototypical}. Denote the center of class $y$ as $\p_y = \frac{1}{K} \sum_{y_i = y} \phi(\x_i)$, then we transform Eq.~\ref{eq:embedding_prediction} to
\begin{equation}
\Pr(\hat{y} = y \mid \x, \mathcal{S},\phi) = \frac{\exp\left(-\mathbf{Dist}(\phi(\x),\; \p_y)\right)}{\sum_{y'\in[N]} \exp\left(-\mathbf{Dist}(\phi(\x),\;\p_{y'})\right)}\label{eq:embedding_prediction2}\;.
\end{equation}

\noindent\textbf{Meta-learning for regression.}
The same embedding-based prediction manner in Eq.~\ref{eq:embedding_prediction} also fits the regression problem, where we have examples $\{(x, y)\}$ in the form of scalar input and  {\em continuous} output pairs:
\begin{align}
&\Pr(\hat{y} = y \mid x, \; \mathcal{S}, \;\phi)\notag\\
&\quad = \sum_{x_i\in\mathcal{S}} \frac{\exp\left(-\mathbf{Dist}(\phi(x),\; \phi(x_i))\right)}{\sum_{x_j\in\mathcal{S}} \exp\left(-\mathbf{Dist}(\phi(x),\;\phi(x_j))\right)} y_i\label{eq:reg_prediction}  \;. 
\end{align}
In Eq.~\ref{eq:reg_prediction}, the prediction of an instance is the weighted sum over the scalar labels of its neighborhood. Similar instances will have close outputs in this way.
Eq.~\ref{eq:reg_prediction} enables meta-learning for regression problems by substituting the predictions into Eq.~\ref{eq:meta-learning2}. We set the loss $\ell$ in Eq.~\ref{eq:meta-learning2} as the mean square error for regression problems. We discuss the classification problem by default in the remaining part of the background and method. Our discussion could be extended to the regression problem easily with Eq.~\ref{eq:reg_prediction}.

\subsection{Contextualized Meta-Learning}
The meta-learned $\phi$ helps classify a query instance given a support set as in Eq.~\ref{eq:embedding_prediction2} and makes visually similar instances close in the embedding space defined by $\phi$.
However, the single $\phi$ neglects the rich semantics of instances, and one instance can be annotated based on various latent attributes.

We assume the label $y_i$ of $\x_i$ can be annotated from $C$ aspects, such as with attributes representing varying levels of granularity\footnote{{In real-world applications, the model is aware of the perspectives used for annotating the instance, \eg, color, category, etc. In other words, $C$ is provided as part of the input.}}.
When we annotate ``red boots'', the label may depend on both ``color'' ($c_1$) or ``category'' ($c_2$) attributes. 
From the ``color'' ($c_1$) view, the label $y_i$ would be assigned as ``red'', but the label $y_i$ becomes ``boots'' when labeling from its ``category'' ($c_2$).
We denote the set of attributes as $\mathcal{C}$. We assume that all examples are annotated based on one of the $C$ latent attributes, \ie, there exists an {\em unknown} $c_i\in[C]$ associated with each $(\x_i, \y_i)$. 
Note that we use labels like ``red'' or ``boots'' for illustrative purposes only, and all labels are transformed into numbers in meta-learning. So we {\em do not have direct access to the attribute labels themselves}.

The ambiguous labeling of instances poses a challenge in determining their relationships. When instances are labeled by different attributes, the relationship between a pair of instances may become contradictory. In such cases, a single $\phi$ is insufficient to capture the diversity of instances.
As illustrated in Fig.~\ref{fig:semantics} (a), the same set of instances may be annotated based on their ``categories'' attribute (\eg, ``Boots vs. Sandals'') in one support set and have labels based on the ``color'' (\eg, ``Red vs. Yellow'') in another support set. 
Furthermore, it is possible to have annotations based on multiple criteria within a single support set $\mathcal{S}$. As depicted in Fig.~\ref{fig:semantics} (b), comparing instances labeled based on different attributes directly, such as "Red" and "3," is not valid.

Due to the difficulty of differentiating the rich semantics among tasks, we propose to {\em contextualize} the meta-learned model $f$. 
Instead of a single $f$ for various kinds of tasks, the contextualized meta-model keeps a set of mapping $\{f_c\}_{c=1}^C$ from the support set to the classifier for these $C$ latent attributes. The context of an instance --- its relationship with other instances in the support set --- facilitates identifying the specific latent attribute $c$ associated with it. Then the meta-model $f_c$ corresponding to the inferred attribute is applied to the support set. 

When we implement the meta-model with embeddings, there are multiple embedding spaces $\{\phi_c\}_{c=1}^C$ in the contextualized meta-model.
We expect instances in an embedding space to only be close if they share similar components {\em measured by the corresponding attribute} and be distant otherwise.
Hence, the corresponding embedding $\phi_c$ helps explain the ambiguous relationship between instances with attribute $c$.
As in Fig.~\ref{fig:semantics}, we define two cases based on the number of attributes involved when annotating instances per task. 

\noindent\textbf{Single Attribute per Task (SAT).}
In SAT, instances in one task (including both the support set $\mathcal{S}$ and the query set $\mathcal{Q}$) share the {\em same} latent attribute $c$.\footnote{We assume that we can sample $N$-way pseudo tasks ($\mathcal{S}$ and $\mathcal{Q}$ pairs) from the base class set $\mathcal{B}$ annotated by every attribute w.l.o.g.}
However, the same instance or visually similar instances in different tasks may have diverse labels due to their different latent attributes.
\begin{itemize}
    \item During meta-training, a contextualized model should discover the latent attribute by taking account of the context $(\mathcal{S}, \mathcal{Q})$. 
    Multiple attribute-specific meta-models or embeddings are learned.
    \item During meta-test, we need to {\em infer} one of the $C$ attributes for a particular task {\em even with novel classes}, \ie, classes non-overlapped with the base class set $\mathcal{B}$.
\end{itemize}
\noindent\textbf{Mixed Attributes per Task (MAT).}
In MAT, assume we sample an $N$-way pseudo task from the base class set $\mathcal{B}$, but the {\em unknown} attribute label $c_i$ and $c_j$ for the $i$-th and $j$-th instances in the same $\mathcal{S}$ or $\mathcal{Q}$ could be different. 
A contextualized meta-model works in a similar manner as confusing supervised learning~\cite{Su2020Task} --- learning over multiple tasks without explicit task labels.
The main challenge is that in MAT, the label of an instance could be annotated based on all latent attributes uniformly, and we cannot infer which attribute to use from the instance directly.
For example, we can equally label a ``red boots'' image as ``red'' (from the ``color'' attribute) or ``boots'' (from the ``category'' attribute) in a sampled task, and both choices are possible given only the image. 
\begin{itemize}
\item During meta-training, an MAT meta-model learns a set of embeddings $\{\phi_c\}_{c=1}^C$ for all attributes in $\mathcal{C}$.
Owing to the previously mentioned challenge, we use the label $y_i$ in addition to $\x_i$ to get more clues for attribute inference. In other words, the ``instance-label'' pair $(\x_i, y_i)$ helps determine the corresponding attribute $c_i$ as well as learn multiple meaningful embedding spaces.
\item  During meta-testing, we may evaluate the contextualized meta-model from two aspects. 
First, we measure the query set performance on the novel tasks to check whether the meta-model is able to construct a more discriminative classifier for a novel support set.
Then, we can also evaluate the quality of multiple learned embeddings. Since we expect the embeddings to reveal the semantic meanings of attributes, we can check the classification ability of the embedding over labels given a specific attribute. We can also visualize the distribution of instances in $\{\phi_c\}_{c=1}^C$ and measure their relatedness to attributes in $\mathcal{C}$.
\end{itemize}

\section{Learning Contextualized Meta-Model}
\label{sec:method}
We propose a general contextualized meta-learning approach {\name} for both SAT and MAT. 
We first summarize the main notion and then describe its configurations. 
\subsection{The Main Idea of {\scshape LeadNet}}
There are two key factors in contextualized meta-learning, including obtaining multiple {\em attribute-specific} meta-models, \ie, the learning strategies from a support set to its target model for various latent attributes, as well as inferring the {\em unknown} attribute for a particular instance.
The ambiguous labeling of an object makes an instance have diverse relationships with others. To capture the polysemous nature when comparing instances, we enrich the semantics by exploring multiple maps, \ie, embedding spaces, that emphasize the diverse similarity relationship w.r.t. various attributes. 
Through modeling the shared properties {\em across} episodes of support and query pairs, we expect those discovered maps to reveal the hidden attributes and the ambiguous comparison relationship could be explained with the help of one or more maps.
{\name} automatically identifies the attribute for relevant instances or a task from two aspects. 
First, it measures the consistency between the similarity of instances measured based on a particular map and the ground truth. Then, {\name} meta-learns to fuse the influences of latent attributes based on the comparison relationships between similar instances.
The learned maps in {\name} facilitate various problems, such as out-of-distribution recognition and few-shot learning.

\begin{figure*}
	\begin{center}
		\includegraphics[width=\textwidth]{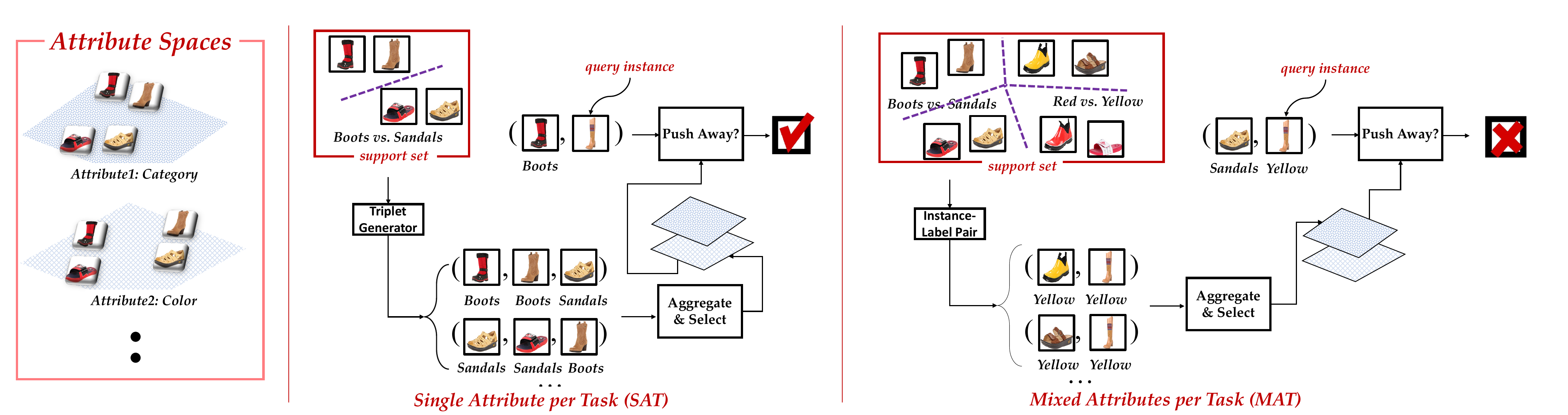}\\
	\end{center}
	\caption{
	{\name} decomposes multiple attributes so that the distance between two instances can be measured from different aspects, and the influence of those comparisons is further fused by taking a holistic view of the multiple attribute spaces (as shown on the left).
	Similar instances w.r.t. a certain attribute are pulled together in their corresponding attribute spaces even if they come from multiple (pseudo) tasks.
	{\it Middle}: In SAT, where there is only one attribute per task, we infer the task-specific attribute with the help of sampled comparison triplets from the support set. 
	{\it Right}: In MAT with mixed attributes per task, the attribute is determined by both the query and the same-class support set instances. Besides, support instances from other classes should be identified whether to push away based on their comparison matching. 
	}\label{fig:method}
\end{figure*}

\subsection{Enriching Semantics via Multi-attribute Discovery}
We generate $C$ attribute spaces with linear maps $\mathcal{L}_C=\{L_1, \ldots, L_C\}$ based on $\phi$. Each $L_c\in\mathbb{R}^{d\times d'}$ emphasizes relevant features in the embedding space. 
The distance in the $c$-th attribute map is $\mathbf{Dist}_{L_c}(\phi(\x),\phi(\x_i)) = \mathbf{Dist}(\phi(\x)L_c, \phi(\x_i)L_c)$. 
We define an instance-specific probability $\kappa_\x\in\mathbb{R}^C$ to measure the similarity between an instance and others w.r.t. the $C$ attributes, whose element $\kappa_\x^c > 0 $ and $\sum_c \kappa_\x^c = 1$. $\kappa_\x$ reveals the ambiguity {\em inside an object}. The more relevant an instance to the $c$-th attribute, the larger the $\kappa_\x^c$, which allocates more probability to attribute $c$ than others.
With a bit of abuse of notation, we still use $\Theta=\{\phi,\mathcal{L}_C,\kappa\}$ to denote all learnable parameters in meta-learning. Given $\kappa$, we reformulate $\Pr(\hat{y} = y \mid \x,\;\mathcal{S},\;\Theta)$ that a query set instance $\x$ similar to the prototype $\p_y$ in the support set 
by integrating the influence of all $C$ attributes:
\begin{equation}
\Pr(\hat{y} = y \mid \x,\;\mathcal{S},\;\Theta) = \;\sum_{c=1}^C {\Pr}_{\x,\p_y}^c\;,\label{eq:leadnetm}
\end{equation}
where
\begin{equation}
{\Pr}_{\x,\p_y}^c=\frac{\exp\left(-\kappa^c_\x\kappa^c_{\p_y}\mathbf{Dist}_{L_c}(\phi(\x),\; \p_y)\right)}{\sum\limits_{y'\in[N]}\sum\limits_{c'=1}^C \exp\left(-\kappa_\x^{c'}\kappa_{\p_{y'}}^{c'}\mathbf{Dist}_{L_{c'}}(\phi(\x),\;\p_{y'})\right)}\label{eq:mmap}\;.
\end{equation}
${\Pr}_{\x,\p_y}^c$ is the attribute-specific posterior probability of the label, which reveals the comparison results of $\x$ and $\p_y$ measured in the $c$-th attribute space.
With Eq.~\ref{eq:mmap}, the prediction of a query instance $\x$ incorporates both the attributes of prototype $\p_y$ and itself in a product form. If both $\x$ and $\p_y$ are relevant to the $c$-th attribute, the product of $\kappa^c_\x$ and $\kappa^c_{\p_y}$ will be large, and their distance computed in the $c$-th attribute space will be mainly considered. 
If none of them possess this particular attribute, $\kappa^c_\x\kappa^c_{\p_y}$ will be small. In this case, their distances in the attribute space do not matter in the posterior integration.

Meaningful local maps could be discovered by learning $\kappa$ for all instances~\cite{Cook2007Visualizing, Maaten2012Visualizing}, which decomposes the semantics of data into multiple views.
To generalize the attribute decomposition ability to novel tasks and construct attribute embedding spaces simultaneously, we meta-learns to decompose an instance in a parametric manner. 
Specifically, assume a mapping $h:\mathbb{R}^d \rightarrow \mathbb{R}^C$ transforms an embedding to a $C$ dimensional vector, which contains sequential fully connected layers and a softmax operator. Then $\kappa_\x=h(\phi(\x))$. Towards contextualizing meta-learning, {\name} maximizes the log-likelihood over $\Theta=\{\phi, \mathcal{L}_C, h\}$:
\begin{equation}
\min_{\Theta} \sum_{(\mathcal{S}, \mathcal{Q})\sim\mathcal{B}}\; \sum_{(\x, y)\in\mathcal{Q}}  -\log\Pr(\hat{y} = y \mid \x, \; \mathcal{S}, \;\Theta)\; +\; \Omega(\Theta)\label{eq:stage1}\;.
\end{equation}
The detailed form of the regularizer $\Omega(\Theta)$ will be discussed at the end of this section. 
Multiple embedding spaces are simultaneously learned during the meta-training progress, corresponding to various attributes in $\mathcal{C}$. 
The discovered multiple maps help explain the similarity between a pair of instances. The prediction of an instance depends on its context --- its relationship with multiple support set instances/prototypes, so the value of the fused distance will vary with selected attributes.
The fused distance in Eq.~\ref{eq:leadnetm}, which weighs and aggregates the distances in every embedding space, makes similar instances in different tasks have close $\kappa$ values. 
Instead of treating same-class instances independently in standard episodic training~\cite{VinyalsBLKW16Matching,Chao2019Meta}, jointly learning their attribute probabilities facilitates the learning experience transition between tasks.

\subsection{Meta Disambiguation}
Eq.~\ref{eq:mmap} utilizes $\kappa^c_\x\kappa^c_{\p_y}$ to reveal the proportion whether to consider attribute $c$ when comparing $\x$ and $\p_y$. 
It makes an independent assumption to simplify the computation, neglecting their joint relevance. 
As in Fig.~\ref{fig:semantics}, single or multiple attributes may exist in a task, and how to annotate instances may vary across tasks.
To this end, we propose to meta-learn a disambiguation module to automatically infer related attributes and aggregate the influence of latent attributes for both SAT and MAT (illustrated in Fig.~\ref{fig:method}). 

\noindent\textbf{Attribute identification for SAT.}
Since there is only one attribute per task, the latent attribute could be inferred by the relationship of all examples in the support set. 
We extract triplets $\mathcal{T}=\{(\x_i, \x_j, \x_l)\}$ from $NK$ instances in $\mathcal{S}$. In a triplet, the target neighbor $\x_j$ is similar to $\x_i$ with the same class, while the impostor $\x_l$ is dissimilar to $\x_i$ (usually $\x_l$ and $\x_i$ have different labels).\footnote{We augment the support set instances to generate target neighbors when $K=1$.}
The in-task comparisons reveal not only the similarity and dissimilarity between instances but also the quality of a local embedding. If a local embedding fits a certain attribute, the similar and dissimilar pairs in the triplet are more obvious than those with other local embeddings. In other words, we can infer the proper latent attribute behind the triplets given the comparison results.

{\name} applies the semi-hard negative policy~\cite{Schroff2015Facenet} to obtain triplets from $\mathcal{S}$ based on $\phi$ of the current optimization stage, and then {\em aggregates} their influences with mapping $\upsilon$:
\begin{equation}
\upsilon(\mathcal{S}) = \mathbf{S}\left(\mathbf{FC}\left(\sum_{(\x_i, \x_j, \x_l) \in \mathcal{T}} \mathbf{FC}\left(\left[\phi(\x_i), \phi(\x_j), \phi(\x_l)\right]\right)\right)\right).\label{eq:SAT}
\end{equation}
$\mathbf{S}(\cdot)$ is the softmax operator, thus $\upsilon(\mathcal{S})$ outputs the probability to choose each of the $C$ attributes. For the triplets sampled from $\mathcal{S}$, embeddings of three elements in the triplets are concatenated, and a fully connected layer with ReLU activation is applied as the transformation. 

The main intuition behind Eq.~\ref{eq:SAT} is that the similarity and dissimilarity among triplets help to infer the specific semantics inside a support set. For example, if ``red boots'' is
similar to ``red sandals'' while dissimilar to ``yellow boots'', the attribute should be ``color''. Then, we can transform the attribute identification problem into learning a mapping from a {\em triplet} to multiple latent attributes, and we can capture the instance order {\em in the triplet} with the concatenation of them~\cite{Flood2017Learning} based on the general embedding $\phi$. 
However, one triplet may not provide a clear preference for a certain attribute, \eg, a triplet with labels (``red sandals'', ``red sandals'', ``yellow boots'') may be valid in both ``color'' and ``category'' attributes. Therefore, we aggregate the choices of multiple triplets together by summing their predictions, which makes the task probability not influenced by the order of those triplets~\cite{Zaheer2017Deep}. The summation in Eq.~\ref{eq:SAT} also reduces the noise in attribute inference and provides better predictions.

Thus, $\upsilon(\mathcal{S})$ encapsulates the relationships among multiple attributes related to $\mathcal{S}$,
which further directs how a particular attribute matters when we predict over $\mathcal{Q}$.
{\name} incorporates $\upsilon$ in $\Theta$ and optimizes Eq.~\ref{eq:stage1} by replacing 
\begin{equation}
\Pr(\hat{y} = y \mid \x, \; \mathcal{S}, \;\Theta)\; = \;\sum_{c=1}^C \upsilon(\mathcal{S})_c\;{\Pr}_{\x,\p_y}^c\;.
\end{equation} 
The $c$-th element $\upsilon(\mathcal{S})_c$ of $\upsilon(\mathcal{S})$ weights the importance of the $c$-th attribute given $\mathcal{S}$. 
We can rethink {\name} as a method to specify the context of a task with compositional attention related to attributes. The attribute weights for each query instance depend on both the instance and task similarity to an attribute. Thus {\name} captures the context of a task with meta-learning and learns to disambiguate the semantics by decomposing attributes for SAT.

\noindent\textbf{Comparison matching for MAT.} When there are mixed attributes in a task, directly comparing instances across multiple learned spaces in Eq.~\ref{eq:mmap} has two issues. 
Similar to SAT, we expect only the distances computed w.r.t. the right attribute to be considered in this case. 
However, unlike SAT, the latent attributes for query instances could vary. Then, the attribute becomes instance-specific, which can only be determined by both the query instance and the same class instances in the support set.
Given a query instance ``red boots'' and a support instance ``red sandals'' both labeled as ``red'', it is easy for us to infer the ``color'' attribute from the $C$ candidates since the two similar instances have visually similar ``colors'' but visually dissimilar ``categories''. 
Thus, observing the similarity and the difference between two similar instances helps infer the unknown attribute.

Moreover, even if a certain attribute is selected, the comparison between one instance with instances from other classes should also be filtered since only the labels annotated by the same attribute can be compared directly. Following the previous example, in the attribute space of ``color'', we only push the instances labeled by ``yellow'' away (with the ``color'' attribute) while neglecting those inter-attribute instances like ``red boots'' labeled as ``boots''.

Since two similar instances could be annotated from two different aspects in MAT, we propose to identify the attribute with both the knowledge of instance and its corresponding labels {\em only during the meta-training progress}. In summary, we also set $\Pr(\hat{y} = y \mid \x, \; \mathcal{S}, \;\Theta)$ equals the weighted sum of attribute-wise logits as in SAT, \ie,
\begin{equation*}
	\Pr(\hat{y} = y \mid \x, \; \mathcal{S}, \;\Theta) = \;\sum_{c=1}^C \upsilon(\mathcal{S}, \x, y)_c \;{\Pr}_{\x,\p_y}^c\;.
\end{equation*}
But the attribute-specific probability ${\Pr}_{\x,\p_y}^c$ is weighted by, 
\begin{align}
&\upsilon(\mathcal{S}, \x, y)_c \label{eq:attribute2}\\
&\quad= \mathbf{S}\left(\mathbf{FC}\left(\sum_{(\x_i, y_i) \in \mathcal{S} \land y_i=y} \mathbf{FC}\left(\left[\phi(\x_i), \phi(\x), \y\right]\right)\right)\right)_c \;.\notag
\end{align}$\upsilon$ selects over $C$ attributes towards an {\em instance-specific} probability. The index $c$ indicates the influence of the c-th attribute. Same-class instance-label pairs differentiate possible attributes. We implement $\mathbf{FC}$ the same as SAT.
Denote $\kappa^c_\x\kappa^c_{\p_y}$ as $\kappa^c_{\x,{\p_y}}$ and $\mathbf{Dist}_{L_c}(\phi(\x), \p_y)$ as $\mathbf{Dist}_c(\phi(\x), \p_y)$, we construct 
\begin{align*}
{\Pr}_{\x,\p_y}^c &=\;\frac{\exp\left(-\kappa^c_{\x,{\p_y}}\mathbf{Dist}_c(\phi(\x), \p_y)\right)}{\sum\limits_{y'\in[N]}\sum\limits_{c'=1}^C \exp\left(-\tau_{\x,\p_{y'}}\cdot\kappa^{c'}_{\x,{\p_{y'}}}\mathbf{Dist}_{c'}(\phi(\x), \p_{y'})\right)}\\
\tau_{\x,\p_{y'}}&=\;\sigma\left(\mathbf{FC}\left([\phi(\x), \;\p_{y'}, \;\y, \;\y']\right)\right).
\end{align*}
$\sigma(\cdot)$ is the sigmoid operator, which squashes the input into [0, 1]. 
$\tau_{\x,\p_{y'}}$ acts as a soft mask that weakens the influence of a particular comparison if $\p_{y'}$ and $\x$ are not in the same attribute space.
With Eq.~\ref{eq:attribute2}, the attribute for each query instance is selected, and the influence of instances annotated by other attributes is automatically ignored --- comparisons in a task are fully contextualized.
Parameters in $\nu$ and $\tau_{\x,\p_{y'}}$ are meta-learned. 

Since the label is only used to help determine the attribute during meta-training, whose effects are similar to a kind of privileged information~\cite{Vapnik2009New,Vapnik2015Learning,David2015Unifying}. Therefore, instances from different attributes are grouped together, and comparisons in the same attribute are leveraged. During the deployment, we evaluate the meta-learned multiple similarity measurements $\phi$ and $\mathcal{L}_C$. We use attribute-specific embeddings to predict an instance from diverse perspectives, or check whether they capture the semantic meaning of the corresponding attribute.

\noindent{\bf Summary of Meta Disambiguation.} Based on the discovered multiple maps, {\name} learns to aggregate the influence of latent attributes, \ie, the similarity-based prediction from multiple maps. By comparing Eq.~\ref{eq:SAT} and Eq.~\ref{eq:attribute2}, we can see that {\name} is general for SAT and MAT with different definitions for the meta-disambiguation module $\nu$. 
In SAT, support and query instances share the same attribute in both the meta-training and meta-test phases. So the ability to infer the latent attribute with $\upsilon(\mathcal{S})$ and weighs multiple learning strategies to obtain a classifier for the query set could be generalized from the meta-training to the meta-test stages.
While in MAT, benefiting from the decomposed $\upsilon(\mathcal{S}, \x, y)$, we directly use the multiple learned embedding spaces ($\phi$ as well as $\mathcal{L}_C$) to analyze the data after meta-training.

\begin{remark}
    {\name} is a general framework to tackle ambiguities, including both SAT and MAT scenarios. There are several works~\cite{tan2019learning,cucurull2019context,mishra2021effectively} addressing the SAT setting from a similar perspective.
In~\cite{tan2019learning}, the similarity condition (\ie, the latent attribute in our work) is directly obtained by the element-wise condition mask. Although it also extracts triplets in model training, these triplets only apply to the loss calculation process and do not influence similarity condition learning. In~\cite{cucurull2019context}, the latent attribute (context) is encoded with graph neural network, while we prove that we can learn suitable context information directly from the triplet comparison. The unsupervised PAN model~\cite{mishra2021effectively} seeks to calculate concept-conditioned similarity via pair-wise comparison. However, it still differs from ours in the way of attribute inference. To summarize, {\name} has inherent differences with these related works, and how we encode attribute information has not been addressed before. Besides, none of these methods achieve compatible performance with {\name} on UT-Zappos in Section~\ref{sec:zappos}, even for PAN with explicit attribute supervision. {\name} performs better than all of them without using attribute information.
\end{remark}

\begin{remark}
Recent years have witnessed the remarkable development of large pre-trained Vision Language Models~(VLMs).
Models like CLIP~\cite{radford2021learning,Paola2022On,Liang2022Mind} not only possess strong recognition ability over a wide range of visual objects, but also have generalization ability with the help of their text encoders to map textual information (such as class names) into the visual classifier. It raises the question of whether we can leverage VLMs to map all attributes to a unified embedding space and address the challenges of both SAT and MAT scenarios.
However, there are several obstacles to applying VLMs in the context of contextualized meta-learning.
First, the latent attributes are unknown in both the meta-training and meta-test phases, which restricts the zero-shot classification ability of VLMs. In addition, as aforementioned, labels are transformed into numerical representations that lack the rich semantic information required for effective encoding by VLMs. Furthermore, our {\name} does not rely on a large-scale pre-training dataset, and could be easily applied to specific problems.
\end{remark}

\begin{figure*}[t]
	\begin{center}
		\includegraphics[width=\textwidth]{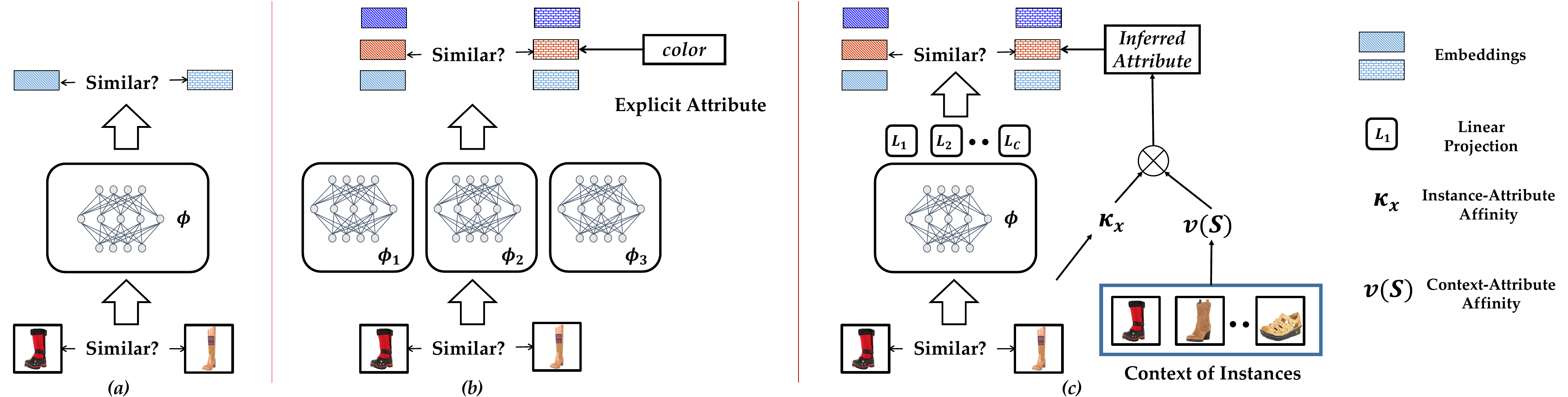}\\
	\end{center}
	\caption{Illustration of different strategies to measure the similarity of a pair of instances, which is a key component in {\name}.
	(a): Traditional methods ignore the hidden attributes behind instances and measure similarities with the same embedding module $\phi$. 
	(b): With explicit attribute supervision, some methods~\cite{mishra2021effectively,veit2017conditional} train more than one embedding to capture the attribute-specific information. The model measures the similarity in the corresponding embedding space.
	(c): Without explicit attribute supervision, {\name} utilizes the context information to automatically infer the proper attribute for a pair of instances. The context-attribute affinity is represented by the triplet (resp. pair-wise) information in SAT (resp. MAT).
	{\name} constructs multiple attribute-specific embeddings with a shared backbone and multiple projections.}\label{fig:structure}
\end{figure*}

\subsection{Regularization of {\scshape LeadNet}}\label{sec:regularization}
The definition of ``attributes'' makes {\name} flexible for various applications. As mentioned in Eq.~\ref{eq:stage1}, $\Omega(\Theta)$ is a regularizer to guide the attribute selection.
Given explicit attributes in advance, \eg, which attribute an instance or a task belongs to, we can set the regularizer $\Omega$ as the cross-entropy between the predicted attribute-wise similarity $\upsilon(\mathcal{S})$ and its ground truth attribute label. Therefore, visually different instances from the same class are pulled together in each attribute space, and key characteristics of the class will be captured to reduce the variance. 

When there are no explicit attribute labels during meta-training, we either rely on the model itself to decompose attributes (as shown in the regression experiments in Section~\ref{sec:regress1}), or we can construct synthetic attributes. 
We use the few-shot classification as an example. We can manually design multiple attributes based on the ``one-vs.-rest'' strategy --- we transform an $N$-way classification task into $N$ one-vs-rest classification sub-tasks. In detail, we treat each base class in $\mathcal{B}$ as a particular attribute, so the prediction of a certain class in a task becomes a weighted sum over predictions over $N$ attributes in $\mathcal{S}$. The semantics in novel classes are also decomposed into the attributes induced by the base classes.
Since the indexes of the $N$ classes are known, we set the regularizer $\Omega$ as in the explicit attribute label scenario during the meta-training. In this way, instances across tasks are also related to the same set of latent attributes (more details are in Section~\ref{sec:benchmark}).

\begin{algorithm}[t]
	\caption{Meta-Training for \name }
	\label{alg1}
	\raggedright
	{\bf Input}: Support set: $\mathcal{S}$; Current model $f(\cdot)$;\\
	{\bf Output}: Updated model; 
	\begin{algorithmic}[1]
		\State Extract triplets from the support set $\mathcal{S}$;
		\For{$(\x_i,\x_j, \x_l)$ in the triplets} 
		\State Calculate attribute-specific embeddings; 
		\State Calculate the instance-specific ambiguity $\kappa_\x$;
		\State Calculate the attribute-specific posterior probability ${\Pr}_{\x,\p_y}^c$ via Eq.~\ref{eq:mmap};
		\State Calculate the meta-disambiguation term $\upsilon(\mathcal{S})$ via Eq.~\ref{eq:SAT} (SAT) or Eq.~\ref{eq:attribute2} (MAT);
		\State Calculate the posterior probability $\Pr(\hat{y} = y \mid \x, \; \mathcal{S}, \;\Theta)$ and update the model;
		\EndFor \\
		\Return the updated model;
	\end{algorithmic}
\end{algorithm}

\subsection{Summary of {\scshape LeadNet}}
{\name} substitutes the posterior probability in Eq.~\ref{eq:mmap} with the compositional ones for SAT and MAT, respectively. 
Then, during the meta-training progress in Eq.~\ref{eq:stage1}, we optimize multiple attribute-specific embedding spaces. To determine the similarity between a pair of instances, we look for the attribute space that satisfies the given similarity the most. If inconsistent similarity exists, the model automatically assigns higher weights to other spaces to minimize the objective, which explores diverse attributes. 
In summary, we contextualize the meta-model by inferring the latent attribute from the support set contexts and applying the corresponding meta-map for similarity calculation.
The ability to make predictions based on a given context and the learned multiple embedding spaces could be used for ``downstream'' tasks. 
We choose the SAT or the MAT variants of {\name} based on the settings of the concrete application, \ie, whether there are one or mixed attributes in a task.
All parameters are meta-learned with the sampled support and query sets jointly.

We summarize the training process of {\name} in Alg~\ref{alg1}.
We also illustrate three typical ways to calculate the similarity for a pair of instances in Fig.~\ref{fig:structure}, which is a key component in {\name}. Traditional methods in (a) do not consider the change of similarity w.r.t. the attribute, and they calculate the similarity with a single embedding $\phi$. Motivated by the diverse similarities under different attributes, some methods~\cite{mishra2021effectively,veit2017conditional} in (b) train multiple embeddings to measure the pair-wise similarity. However, they rely on explicit attribute information to select the right embedding space. In this paper, we propose to meta-learn the strategy to infer the attribute via the context of instances. Therefore, {\name} will automatically choose the proper attribute as well as the embedding space without explicit supervision.

\section{Related Work}
\label{sec:related}
\noindent{\bf Meta-learning} emphasizes the generalization of the learned inductive bias to {\em novel tasks} by sampling (pseudo) tasks from the base class data, which mimics the future deployment environment~\cite{VinyalsBLKW16Matching,FinnAL17Model}.
The inductive bias, \ie, the experience of learning the ``learning strategy from the support set to the target model'', is shared across various tasks. Thus, we can construct a classifier efficiently once with a support set, and that classifier predicts all same-distribution query set instances well.
Initialization of deep neural network~\cite{Nichol2018On,Khodadadeh2019Unsupervised}, instance embeddings~\cite{SnellSZ17Prototypical,Metz2018Learning,Flood2017Learning,Scott2018Adapted,Lee2019Meta,fei2021z,ye2022revisiting}, optimization policies~\cite{ravi2016optimization,Lee2019Meta,ye2020few,baik2021meta,wang2022global}, and data augmentation~\cite{Wang2018Low,Li2019Learning} could be meta-learned across sampled tasks towards an effective ``learning strategy''.
For example, we achieve an effective classifier based on the labels of neighbors in the meta-learned embedding space, or by fine-tuning a meta-learned model initialization with fixed steps.
Meta-learning has become the key technique in applications such as few-shot learning~\cite{ye2020fewshot,ye2020heterogeneous,tian2022improving,wei2022data}, recommendation system~\cite{vartak2017meta,luo2020metaselector}, and robot navigation~\cite{wortsman2019learning,gupta2018unsupervised}. 
In few-shot learning, \ie, there are only limited support examples, so the meta-learned ``learning strategy'' becomes helpful. 

\begin{figure*}[t]
	\centering
	\begin{minipage}{0.19\linewidth}
		\includegraphics[width=\linewidth]{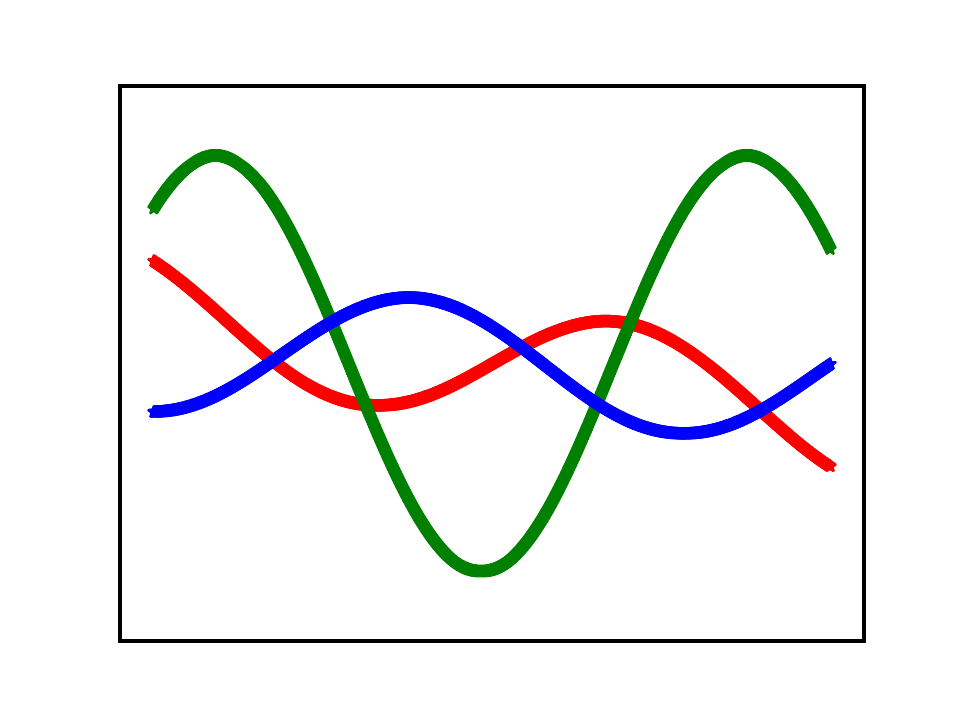}
		\centering\mbox{(a) Ground Truth}
	\end{minipage}
	\hfill
	\begin{minipage}{0.19\linewidth}
		\includegraphics[width=\linewidth]{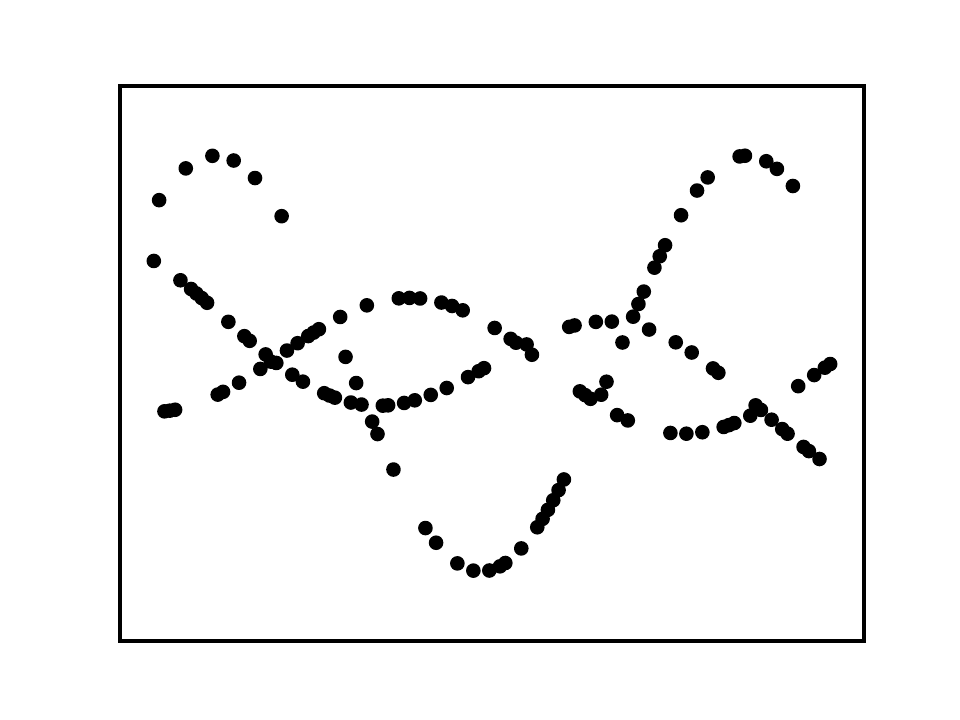}
		\centering\mbox{(b) $\mathcal{S}$ w/o Attribute Labels}	
	\end{minipage}
	\hfill
	\begin{minipage}{0.19\linewidth}
		\includegraphics[width=\linewidth]{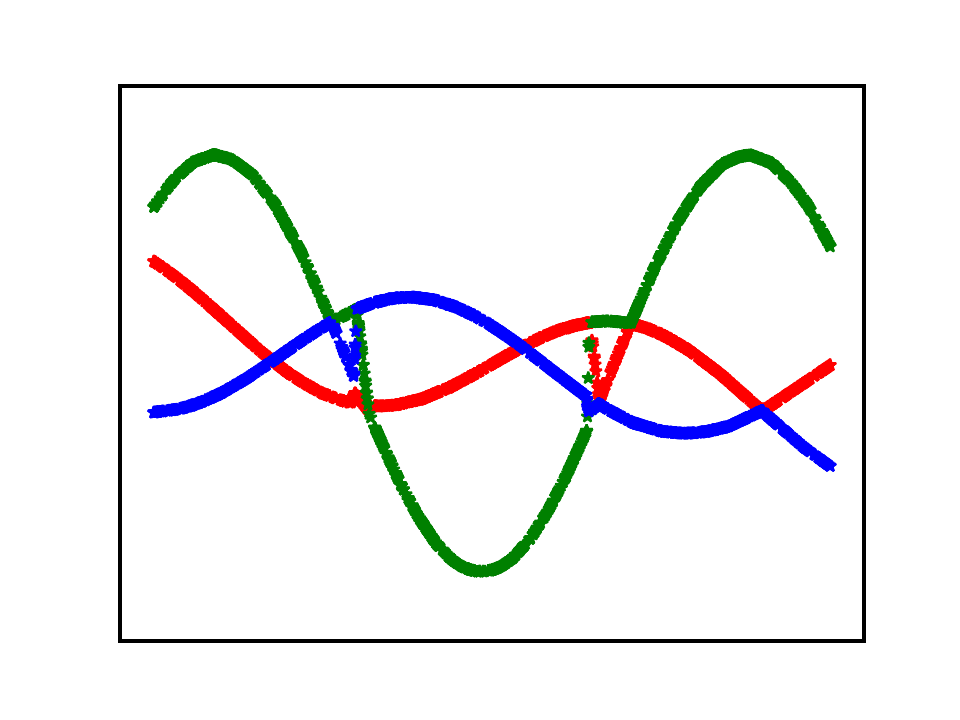}
		\centering\mbox{(c) CSL}	
	\end{minipage}
	\begin{minipage}{0.19\linewidth}
		\includegraphics[width=\linewidth]{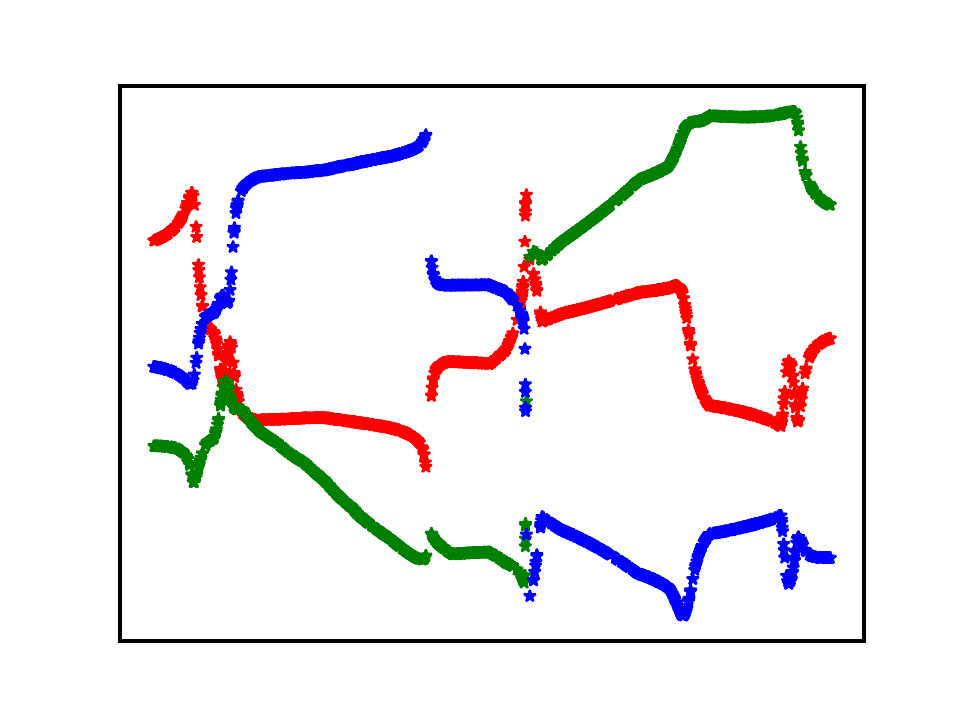}
		\centering\mbox{(d) {\name}$^-$}	
	\end{minipage}
	\begin{minipage}{0.19\linewidth}
		\includegraphics[width=\linewidth]{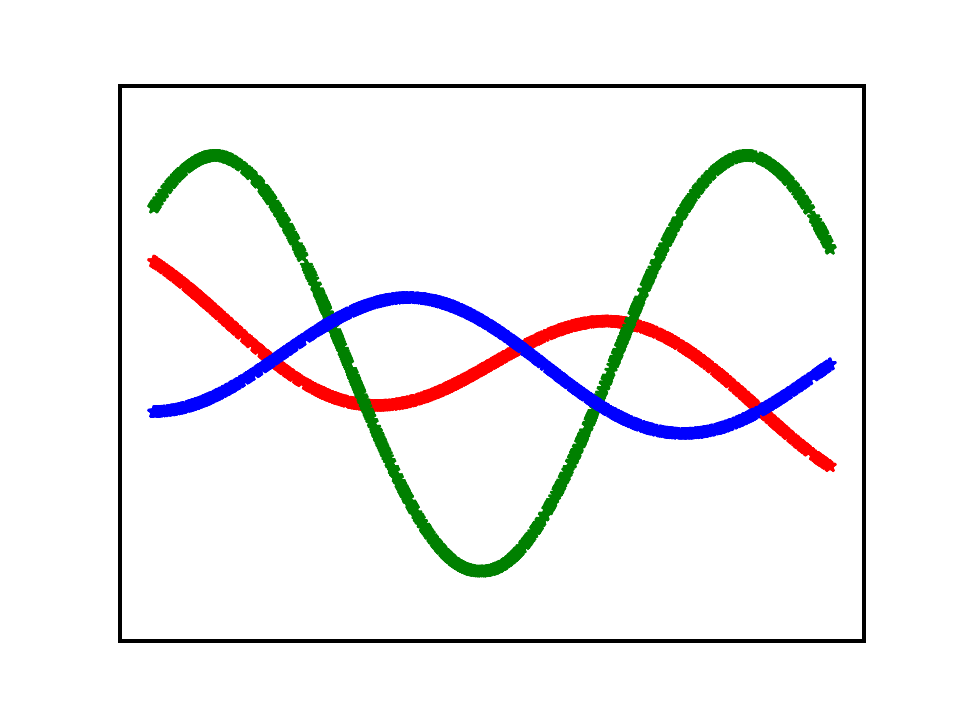}
		\centering\mbox{(e) {\name}}	
	\end{minipage}
	\caption{Visualization of the confusing regression problem. (a) There are three latent attributes (three curves plotted with different colors) in the training data, and one instance could be labeled by the output from a randomly selected curve. (b) The training set without attribute labels. 
	(c)-(e): the predictions of three attributes by CSL, {\name}$^-$, and {\name}, respectively. {\name}$^-$ is the degeneration of {\name} without meta-disambiguation module.
	In the confusing regression problem, we could not infer the latent attribute only given $x$, since an instance will be labeled by the three curves equally. Thus CSL and {\name} incorporate $(x, y)$ to identify the attribute during model training. 
	} \label{fig:confusing_reg}
\end{figure*}

To resolve the issue that a uniform ``learning strategy'' in vanilla meta-learning can not deal with the diversity among (pseudo) tasks, some methods~\cite{ye2020few,Zhang_2020_CVPR,ye2020fewshot} propose to meta-learn task-specific initialization or transformation. However, they are not designed for the scenario where instances are labeled from attribute to attribute across the sampled (pseudo) tasks. In other words, the context of an instance is not fully captured.
{\name} proposes to contextualize meta-learning, which learns to discover latent attributes among tasks and the strategy to weigh/select them. 
The learned multiple latent attributes in {\name} also facilitate the similarity measurement in few-shot learning, which achieves better performance on benchmarks.

\noindent{\bf Multi-task learning} is similar to meta-learning, which extracts inductive bias by learning over multiple tasks simultaneously~\cite{Argyriou2006Multi,Maurer06Rademacher,Maurer2008Bounds,Maurer09Transfer}. The inductive bias, such as embedding and partial top-layer classifiers~\cite{Wang2016Multiplicative,Ma2018Model,Shiarlis2018TACO,Sun2020Ada} generalizes to instances sampled from one of the multiple task distributions. 
Multi-task learning leads to a more compact and generalizable model for discerning over {\em novel instances} from those tasks~\cite{MaurerPR16The}, which has been wildly used in clustering~\cite{Zhang2015Convex}, face recognition~\cite{Huang2021When}, anomaly detection~\cite{Georgescu2021Anomaly}, etc. 
In standard multi-task learning, the correspondence between an example with one of the multiple tasks is explicit~\cite{Lin2019Pareto,Standley2020Which,Chen2020Just}, so we can differentiate the task-specific characteristic easily.
Invariant risk minimization~\cite{Arjovsky2019IRM,Choe2020An,KamathTSS21Does,RosenfeldRR21Risk} takes advantage of the shared property over multiple tasks to determine the invariant features for out-of-distribution recognition.
However, due to the ambiguity of latent semantics, the task label sometimes is unknown during training, and the model is asked to consider multiple latent attributes in this confusing scenario~\cite{Su2020Task,CreagerJZ21Environment}. 
Thus, in multi-task learning without explicit task labels, a model is required to infer from the latent attributes of various tasks, and standard multi-task learning methods cannot perform directly. 
{\name} learns over multiple (pseudo) tasks to identify the rich components from the base class data towards discovering latent components in meta-learning. The ability to deal with MAT enables {\name} in confusing multi-task learning without task labels.

\noindent{\bf Learning with rich semantics.} 
Since instance representation is challenging to cover multiple aspects of complicated data, taking account of the diverse characteristics of instances is important, and has been found useful in QA~\cite{nie2020dc,le2020hierarchical}, fashion compatibility recommendation~\cite{cucurull2019context,lin2020fashion,yang2019interpretable,guan2022partially, lu2022learning}, retrieval~\cite{zhang2019pcan,tamine2018evaluation} and scene understanding~\cite{yu2020context,dvornik2019importance}.
Since a single feature space or the common initialization could not reveal the multiple aspects of data, to explore the rich semantics~\cite{plummer2018conditional,Zhao2018Modulation,ahmad2022deep,zhou2022cross}, non-euclidean relationship~\cite{changpinyo2013similarity,amid2015multiview} and conditional similarity~\cite{Nigam2019Towards,tan2019learning,veit2017conditional,ye2022identifying} are investigated. 
Instance pairs or triplets are hard to resolve the relationship ambiguity and determine the latent attribute, which usually relies on the explicit attribute labels~\cite{veit2017conditional,yang2019interpretable,mishra2021effectively}.
For example, \cite{Tokmakov2019Learning,Xing_2019_NIPS,Ren2021Wandering} compose instance representations with the supervision of attributes.

To discover latent attributes, \cite{tan2019learning} takes advantage of pre-sampled attribute-specific triplets, and~\cite{Cao2021Concept} utilizes the image patches when mining fine-grained visual concepts. 
\cite{Scott2018Adapted,Vuorio2019Multimodal,Zintgraf2019Fast,Yao2019Hierarchically,ye2020fewshot} adapts the feature embeddings or the model initialization given a few-shot support set with meta-learning.
{\name} is a general meta-learning approach where there has one (SAT) or multiple (MAT) attributes in a sampled (pseudo) task without the explicit supervision of attribute information.
Through contextualizing a meta-model, {\name} learns to fuse the effects of multiple attributes over both instances as well as tasks and adapts the embeddings based on attributes. Thus, semantics related to attributes are fully explored in those learned spaces.

\noindent{\bf Learning with multiple labels.} 
In many real-world problems, instances are annotated based on various aspects and may have multiple labels~\cite{ZhangZ14,AlfassyKASHFGB19}. Recent studies have extended the concept of multi-label learning to the few-shot scenario, where meta-learning has emerged as a powerful tool~\cite{wu2019learning,ChenLCHW22,simon2022meta}. However, in standard multi-label learning, instances are aware of all related labels and have explicit annotations.
In contrast, in our contextualized meta-learning, although there are multiple attributes associated with instances, the specific attribute associated with each instance is unknown. It is the meta-model's duty to infer the appropriate attribute. {\name} addresses this challenge by learning multiple embedding spaces that consider the context of the support set.
In our experiments (please refer to Table~\ref{tab:zappos}), we demonstrate that directly applying strategies from multi-label meta-learning to our contextualized meta-learning tasks leads to inferior performance.
\section{Experiments}
\label{sec:experiment}

We verify the ability of {\name} to contextualize the model in various fields, namely, 
\begin{itemize}
	\item Confusing regression with multiple components (Section~\ref{sec:regress1});
	\item Ambiguous few-shot classification on UT-Zappos50k (Section~\ref{sec:zappos});
	\item Debiasing few-shot classification on Colored-Kuzushiji (Section~\ref{sec:ood});
	\item Confusing classification on colored digits (Section~\ref{sec:confusing_cls});
	\item Benchmark few-shot classification tasks on {\it Mini}ImageNet / {\it Tiered}ImageNet / CUB, as well as the cross-domain few-shot classification (Section~\ref{sec:benchmark}).
\end{itemize}
We use the {\bf MAT variant} of {\name} for Section~\ref{sec:regress1} and~\ref{sec:confusing_cls} and investigate {\bf SAT variant} of {\name} for the other problems. {\bf No explicit attribute labels are provided in our experiments}. For each problem, we provide its background, settings, comparison methods, implementation details, evaluation criterion, and results.
Unless specifically stated, all the compared methods utilize the same network backbone, optimizer, learning rate scheduler, batch size, and training epochs. We refer to the original paper to set the hyper-parameters of compared methods.
All experiments are deployed with Nvidia 2080Ti. The source code of {\name} is available at \url{https://github.com/zhoudw-zdw/TPAMI-LeadNet}.

\subsection{Regression with Confusing Components} \label{sec:regress1}
Different from the standard regression where a scalar $x$ is mapped to a single continuous value $y$, here we consider the ambiguous regression task and associate one $x$ with {\em multiple} possible scalar outputs. 
In other words, there can be many underlying curves to label an input, and our {\name} should also identify those latent attributes and learn multiple regressors for them.
This {\bf MAT} scenario is the same as the confusing supervised regression~\cite{Su2020Task} --- learning with multiple regression tasks without task labels.

\noindent{\bf Setups.} There are three latent attributes in a regression task, we follow~\cite{Su2020Task} and link attributes with three sine curves, \ie, $y_1 = -0.4 x + 0.9\sin(2x)$, $y_2 = -3\sin(2x + 1.7)$, $y_3=-0.1x+0.9\sin(2x + 2.5)$.
Given one $x$, we sample one of the three functions (say the second one) and then set its label based on the latent output of the curve ($y_2(x)$ in this case). 
Directly learning with the confusing data without annotations of attribute labels will fit the average value and could not differentiate these attributes. 
We compare different methods by making predictions on uniformly sampled instances and compare the predicted curves with the ground truth. A good method predicts the values of three latent curves well and infers the latent attributes correctly.

\noindent{\bf Implementations.} In the regression problem, we make prediction with Eq.~\ref{eq:reg_prediction} and use square loss in Eq.~\ref{eq:meta-learning2}.
We implement $\phi$ as a three-layer fully connected network with a hidden layer of 32 dimensions. Since all labels are continuous, we combine the instance and label and implement $\upsilon$ with a two-layer fully connected network during meta-training. All methods are optimized with Adam~\cite{KingmaB14ADAM} using an initial learning rate of 0.1 over 50 epochs. The cosine annealing learning rate schedule is used. We set both the support and query set with size 128. Although enough instances should be collected to infer the latent attribute, {\name} does not need a very large batch size in the implementation, which is sampling efficient. We also utilize a few-shot initialization as~\cite{Su2020Task}. In detail, we have ten randomly selected examples and all their mapped outputs from three functions. The limited ground truth examples help infer the latent attributes in training.
Given learned {\name}, we infer the latent attribute of each training example $(x, y)$ based on $\nu(\mathcal{S}, x, y)$, so that we obtain attribute-specific support sets from the whole training set. Then we can predict a novel instance for the $c$-th attribute as Eq.~\ref{eq:reg_prediction} with $\phi$ and $L_c$.

\noindent{\bf Results.} 
Results are in Fig.~\ref{fig:confusing_reg}. There are three latent curves denoted by different colors in (a). The model is trained with the data in (b), where \emph{no attribute labels are provided}. The model should infer which attribute an instance is labeled via the example itself and the relationship with other instances in the context. 
To stress the importance of meta-disambiguation in {\name}, we construct a degeneration variant {\name}$^-$, which predicts with Eq.~\ref{eq:leadnetm}. Without the meta-disambiguation module $\nu$, {\name}$^-$ is hard to train and to differentiate attributes as in (d). 
We mainly compare our model with the recently proposed CSL~\cite{Su2020Task}. The results of CSL in (c) can identify the latent attributes successfully. However, CSL mixes up instances from different attributes near the intersection point of functions, and the predicted curves are not smooth.
In {\name}, we use three projections to differentiate the attributes, which is more parameter-efficient than CSL. Besides, we do not need alternative optimization as CSL. The results of {\name} are shown in (e). {\name} can recover all latent attributes, and the curves are close to the ground truth. The results validate the ability of {\name} to contextualize a model.

\subsection{Ambiguous Few-Shot Classification} \label{sec:zappos}
{\name} is able to identify the attribute from an ambiguous environment. 
Considering a real-world scenario where an object can be described by different attributes with diverse labels. When the contexts are generated by selecting a set of objects as the support set under a certain attribute space, an ideal model should tell the picked attribute and predict corresponding query set labels correctly, without the supervision of attribute labels.

\noindent{\bf Setups.} 
We test {\name} on UT Zappos50K~\cite{Yu2017Semantic,Li2019Repair}, where one image could be measured from six different attributes, namely, closure, heel height, insole, material, subcategory, and toe style. Based on a particular view of data, all instances are categorized differently. There are 16 categories on average over all attributes. 
We complete a few missing values in the dataset with their corresponding major attributes.
For each task, one of the six attributes is selected, and instances are labeled with the specified attribute. Therefore, instances in a task belong to one attribute, labeled based on the same perspective.
Due to the ambiguity, a uniform model cannot cover all attributes. The latent attribute of a task will only be identified based on similar and dissimilar objects. 
We evaluate the performance by sampling 1-shot 5-way and 5-shot 5-way tasks (with 15 query examples per class). We check whether the model could infer the latent attribute related to the task and make predictions well.

\noindent{\bf Implementations.}
We use the ResNet-12~\cite{he2016deep, Lee2019Meta} to implement the model encoder $\phi$. Besides, we implement $h$ as a three-layer fully connected network with a hidden layer of 1280 dimensions. {\name} does not use attribute labels. We follow~\cite{ye2019learning} to set the hyper-parameters.

\begin{figure}[t]
	\centering
	\begin{minipage}{0.49\linewidth}
		\includegraphics[width=\linewidth]{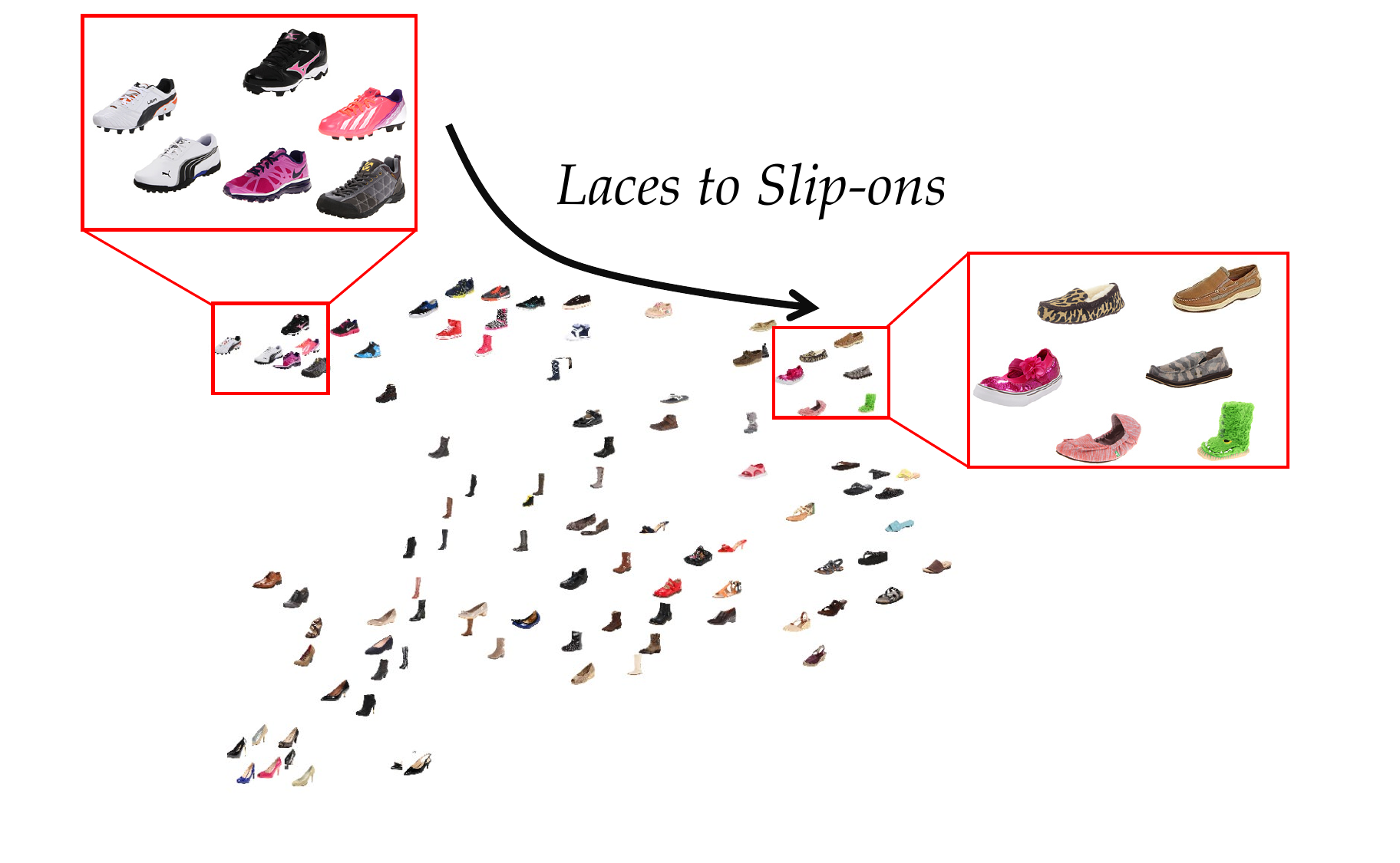}\label{fig:laces}
		\centering\mbox{Laces}
	\end{minipage}
	\hfill
	\begin{minipage}{0.49\linewidth}
		\includegraphics[width=\linewidth]{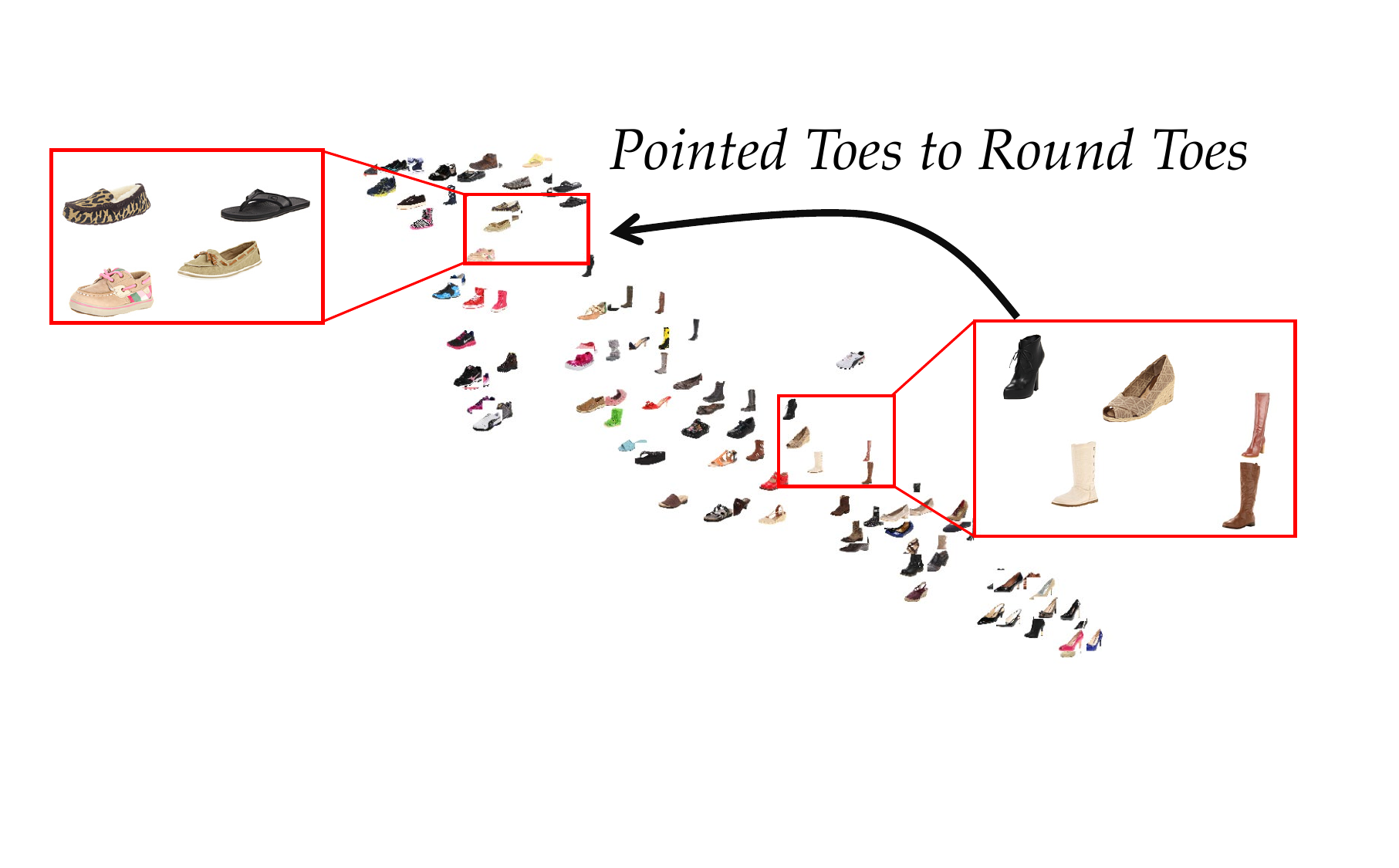}\label{fig:toestyle}
		\centering\mbox{Toe style}	
	\end{minipage}
	\caption{Additional t-SNE embeddings for learned attributes of {\name} w.r.t. Fig.~\ref{fig:semantics} (a). Attribute learns to differentiate shoes based on laces (left) and Toe style (right), respectively.} \label{fig:tsne}
\end{figure}

\begin{table}[t]
	\centering
	\caption{Averaged 1/5-Shot 5-Way classification results on UT-Zappos50K dataset over 2000 trials. Methods with $\dagger$ are trained with explicit attribute labels, while others are trained without them. {\name} outperforms the methods with attribute labels.
	}\label{tab:zappos}
	\begin{tabular}[t!]{l|cc}
		\addlinespace
		\toprule
		Setups $\rightarrow$ &1-Shot 5-Way & 5-Shot 5-Way \\
		\midrule
		SCE-Net{~\cite{tan2019learning}} & 32.60$\pm$0.84 &  43.71$\pm$0.91  \\
		CGAE{~\cite{cucurull2019context}} & 31.25$\pm$0.98 & 41.06$\pm$0.86 \\
		CSA-Net{~\cite{lin2020fashion}} & 35.22$\pm$0.84 & 44.13$\pm$0.77 \\
		PAN-Unsupervised{~\cite{mishra2021effectively}} & 34.15$\pm$0.82 &  44.57$\pm$0.71  \\
		PAN-Supervised$^{\dagger}${~\cite{mishra2021effectively}} & 38.83$\pm$0.61 &  46.92$\pm$0.60  \\
		ProtoNet{~\cite{SnellSZ17Prototypical}} & 38.19$\pm$0.55& 46.87$\pm$0.60 \\ 
		RelationNet+NLC{~\cite{simon2022meta}} & 35.41$\pm$0.60 & 44.18$\pm$0.56 \\
		Castle{~\cite{ye2019learning}} & 38.46$\pm$0.55 & 45.86$\pm$0.56 \\ 
		MetaOptNet{~\cite{Lee2019Meta}} & 38.66$\pm$0.55 & 46.55$\pm$0.58 \\ 
		MuMoMAML{~\cite{Vuorio2019Multimodal}} & 35.75$\pm$0.92  &43.04$\pm$0.66\\ 
		TADAM{~\cite{Oreshkin2018TADAM}} & 37.02$\pm$0.63  &44.88$\pm$0.71\\ 
		ProtoAttribute$^{\dagger}$  & 38.75$\pm$0.57 & 47.18$\pm$0.60 \\ 
		\midrule
		{\name} &  \bf 39.19$\pm$0.57 & \bf 48.29$\pm$0.58 \\  
		\bottomrule
	\end{tabular}
\end{table}

\noindent{\bf Results.}
We show t-SNE visualizations of four attributes learned by {\name} in Fig.~\ref{fig:semantics} (a) and Fig.~\ref{fig:tsne}. It shows that by learning to decompose the task with hidden attributes, \name~can learn to differentiate instances from different perspectives. For example, in Fig.~\ref{fig:tsne} (left), the model embeds instances from the perspective of having laces. Shoes on the left-hand side are those with laces, while shoes on the right-hand side are slip-ons. Fig.~\ref{fig:tsne} (right) illustrates a similar scenario that shoes on the left-hand side are round-toes, while shoes on the right-hand side are pointed-toes.

We compare with various baselines like MetaOptNet~\cite{Lee2019Meta}, PAN~\cite{mishra2021effectively} and SCE-Net~\cite{tan2019learning}, and Table~\ref{tab:zappos} records the few-shot classification results.
Note that most methods (including ours) in Table~\ref{tab:zappos} do not use attribute labels except those with $\dagger$, \ie, PAN-Supervised, and ProtoAttribute.
Some task adaptation methods like MuMoMAML~\cite{Vuorio2019Multimodal}, TADAM~\cite{Oreshkin2018TADAM}, and Castle~\cite{ye2019learning} perform worse than MetaOptNet, since using instances in the few-shot support set independently could not determine selected attributes correctly. SCE-Net needs pre-collected triplets to learn the relationships among tasks, and a high-quality pre-trained embedding is necessary to achieve satisfactory performance. PAN improves SCE-Net with the ability to utilize attribute labels, which gets better performance than SCE-Net. Training with explicit attribute labels, PAN-Supervised improves the accuracy than PAN-Unsupervised, but {\name} outperforms them even without the attribute supervision. Additionally, we find multi-label meta-learning method RelationNet+NLC performs poorly, verifying that the ability to infer the hidden attribute is essential in this setting.
We also construct a reference method ProtoAttribute, which utilizes the task attribute label during the meta-training. Multiple ProtoNet models are learned for each attribute in ProtoAttribute, and their predictions are averaged at last. 
By comparing ProtoAttribute with ProtoNet~\cite{SnellSZ17Prototypical}, we find the supervision of the specified hidden attribute helps to improve ambiguous classification performance.
{\name} gets better accuracy in both cases, validating the effectiveness of the comparison-driven selector to identify the context of a task.

\begin{figure}[t]
	\centering
	\includegraphics[width=0.5\textwidth]{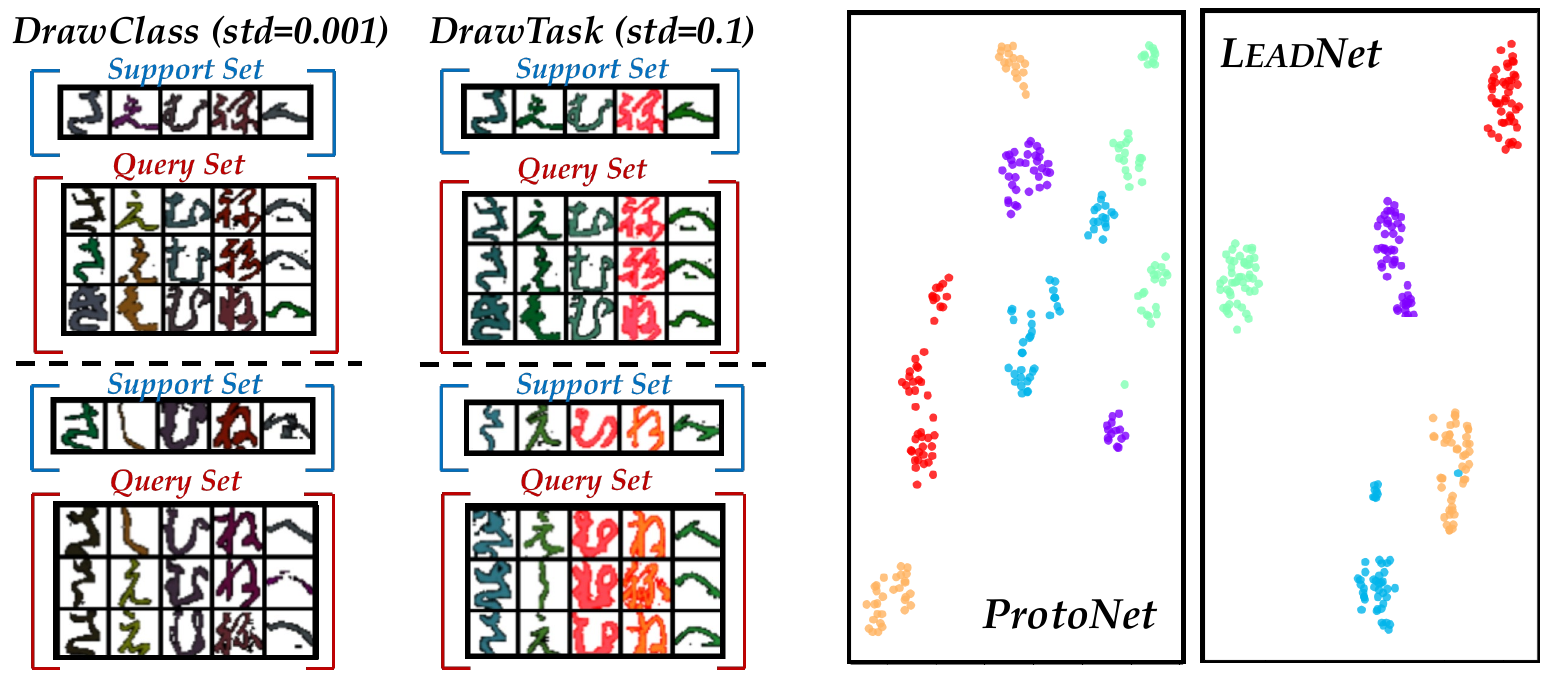}\\
	\caption{Illustrations of two drawing plans on few-shot classification tasks on Colored-Kuzushiji. The leftmost two columns correspond to ``DrawClass'' and ``DrawTask'' plans. A robust model will not be disturbed by the perturbation ``color'' attribute, so {\name} learns more compact embeddings than ProtoNet. We show the t-SNE of embeddings over DrawClass'' tasks in the right plot.}\label{fig:ood}
\end{figure}

\subsection{Debiasing Few-Shot Classification} \label{sec:ood}
In this section, we test whether {\name} will be influenced by the disturbed semantics in the data and whether the contextualized model is robust w.r.t. the out-of-distribution~(OOD) or biased tasks. 
We expect that by figuring out the latent attributes, {\name} is able to learn embeddings invariant to the change of the context.

\begin{table}[t]
	\caption{1-Shot 5-Way classification results on Colored-Kuzushiji with two drawing plans, \ie,  ``DrawClass'' and ``DrawTask''. ``std'' denotes the perturbated noise over colors.}\label{tab:ood}
	\centering
	\begin{tabular}{c|cc}
		\addlinespace
		\toprule
		DrawClass & std=0.001 & std=0.1 \\
		\midrule
		ProtoNet{ ~\cite{SnellSZ17Prototypical}} & 77.13$\pm$0.85 & 94.26$\pm$0.43 \\
		MuMoMAML{ ~\cite{Vuorio2019Multimodal}} & 77.14$\pm$0.83 & 93.34$\pm$0.42 \\
		MetaOptNet{ ~\cite{Lee2019Meta}} & 79.62$\pm$0.83 &  90.28$\pm$0.59 \\
		\midrule
		{\name} & \bf 84.43$\pm$0.79 & \bf 97.32$\pm$0.27 \\
		\bottomrule
		\toprule
		DrawTask & std=0.001 & std=0.1 \\
		\midrule
		ProtoNet{ ~\cite{SnellSZ17Prototypical}} & 94.87$\pm$0.51 & 92.34$\pm$0.61 \\
		MuMoMAML{ ~\cite{Vuorio2019Multimodal}} & 97.64$\pm$0.33  & 94.26$\pm$0.34 \\
		MetaOptNet{ ~\cite{Lee2019Meta}} & 98.80$\pm$0.27 & 98.53$\pm$0.28 \\
		\midrule
		{\name} & \bf 99.76$\pm$0.10 & \bf 99.60$\pm$0.16 \\
		\bottomrule
	\end{tabular}
\end{table}

\noindent{\bf Setups.}
Similar to~\cite{Arjovsky2019IRM,Kim2019Learning,Li2019Repair}, which plot RGB colors over the grayscale images based on the Kuzushiji-49 dataset~\cite{Clanuwat2018Deep}, and test whether a model will be disturbed by the colors.
Kuzushiji-49 contains 270,912 grayscale images with size 28 by 28 spanning 49 classes, including 48 Hiragana characters and one Hiragana iteration mark. 
To increase the difficulty of the character classification problem, we mix each image with the disturbed ``color'' attributes by drawing colors on the grayscale images. 
Then a model will be confused if it makes predictions on the character shape and color jointly and will fail to predict the character when colors change. 
Assume each class is drawn with three different colors. For instance, in a task, we first randomly choose one of the three candidate colors and then add a perturbation mean zero noise over the selected color before drawing the image. The larger the standard deviation of the noise, the more a selected color will be further perturbed. The two plans differ in their drawing manners:
\begin{itemize}
	\item ``DrawClass'' selects a color from the class-specific candidate sets for every image and treats all instances independently. Therefore, in a task, two instances from the same class will have different colors. 
	\item ``DrawTask'' keeps the same class instances from both support and query sets selecting the same candidate color and only varies the selected color across tasks.
\end{itemize}
For the ``DrawClass'' plan, a few-shot task is more complicated if the strength of the color noise is small, while for ``DrawTask'', a task becomes difficult with larger std values. 

We split the dataset and randomly choose 30 classes as the base class, and the remaining classes are the novel ones (with 1000 images per class). 
Non-overlapping colors are used during meta-training/test stages. For those base classes, we select RGB values in [0, 0.5], while for novel classes, we set RGB values in [0.5, 1].
The 1-shot 5-way classification results over 1000 trials are evaluated, and the mean accuracy and 95\% confidence interval are reported.

\noindent{\bf Implementations.}
For all models, we implement the backbone with a four-layer convolution network~\cite{VinyalsBLKW16Matching,SnellSZ17Prototypical}. In each block, there is a sequence of convolution, batch normalization~\cite{ioffe2015batch}, ReLU, and max pooling operators. Setting the hidden dimensions to 64, we add a global average pooling at the last layer. For MuMoMAML~\cite{Vuorio2019Multimodal}, we adopt its more complicated backbone.

\noindent{\bf Results.}
Fig.~\ref{fig:ood} illustrates the colored tasks with two plans, and the 1-shot 5-way classification results over 1000 trials are listed in Table~\ref{tab:ood}. 
For the ``DrawClass'' plan, a few-shot task is more difficult if the strength of the color noise is small, while for ``DrawTask'', a task becomes difficult with larger std values. 
In addition to the ``shape'' attribute for characters corresponding to the category labels, the ``color'' attribute acts as a disturbance. Direct learning a model (\eg, ProtoNet) is prone to fit the color attribute in a few-shot task, so it will fail to recognize novel classes. A robust model could handle the biased scenario well. It will focus on the characteristic shape to construct a classifier for each support set and generalize to novel tasks with other colors.
MetaOptNet works better than ProtoNet, which is consistent with the intuition in~\cite{Arjovsky2019IRM}. {\name} achieves the best results in both cases since it relates classes in different tasks so that it will remove the disturbing colors and build robust models in out-of-distribution/biased scenarios.

\subsection{Confusing Classification on Colored Digits}\label{sec:confusing_cls}
Since one instance could be annotated from different perspectives, we consider the confusing classification problem where examples may belong to different latent attributes in a task. Directly learning over all labels will neglect the relationship between classes in an attribute and miscalibrate the classes across attributes. 
We construct a multi-task classification problem without task labels --- which latent task an example comes from is unknown. The {\bf MAT} model needs to identify the latent attributes, \ie, attribute labels, and construct classifiers for all attributes.

\begin{figure}[t]
	\centering
	\includegraphics[width=0.9\linewidth]{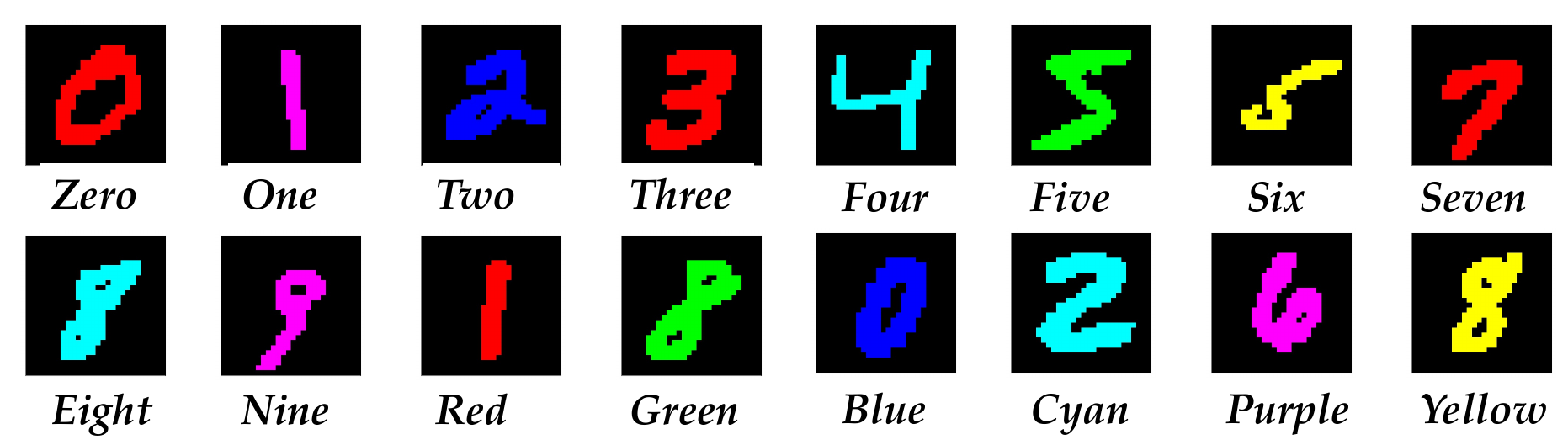}
	\caption{Samples of training data in the confusing classification task, which contains two latent attributes. An image could be labeled from either ``color'' or ``digit'' aspects.} \label{fig:confusing_cls}
\end{figure}

\noindent{\bf Setups.}
We follow~\cite{Su2020Task} and extend MNIST~\cite{Lecun1998MNIST} by adding six random colors for all images.\footnote{We follow the code of~\cite{Su2020Task}, which is a bit different from the setting in their paper. For example, the code only provides the seed to generate 6 (instead of 8) random colors.} 
There are ten classes with digits 0-9 in the dataset. We add six random colors for all images so that each image could be annotated by its color or digit from the 16 candidate labels. Some sampled images from the dataset can be found in Fig.~\ref{fig:confusing_cls}. For example, the first image is a ``red zero'', but we label it from the digit perspective as ``zero''. Therefore, there are two latent tasks in this dataset, \ie, color classification and number classification. We use the same training-test splits as~\cite{Lecun1998MNIST}.
A model is required to classify both color and digit tasks corresponding to two latent attributes. 
As mentioned before, the MAT model utilizes both instances and labels in a sampled task only during training to determine the instance-specific latent attribute, so that latent classifiers target two attributes are learned accordingly.
We evaluate the model for both attributes separately by the averaged accuracy per attribute as~\cite{Su2020Task}.
For a test instance with two conceptual labels, we compute the classification ability of the model upon each attribute, respectively.
To take account of the exchange equivalence between attributes, we search the latent correspondence between multiple learned classifiers and attributes based on the attribute-wise accuracy\footnote{After the meta-training process, we evaluate the accuracy on both attribute spaces separately, similar to the traditional metric learning process. Since the meta-training and meta-testing process tackles the same class space, there are no support instances during meta-testing.}.
The vanilla supervised learning baseline SL directly learns a 16-way classifier and outputs the class with the largest logit. 

\begin{table}[t]
	\centering
	\caption{Confusing classification performance over colored MNIST. The average accuracy over color and digit attributes are  reported.}
	\begin{tabular}{l|cc}
		\addlinespace
		\toprule
		Setups $\rightarrow$ & Color & Digit \\
		\midrule
		SL    & 45.87   & 49.25 \\
		CSL{~\cite{Su2020Task}}        
		&  94.31   & 61.60 \\
		\midrule
		{\name} &  \bf 95.39 & \bf 70.89 \\
		\bottomrule
	\end{tabular}
	\label{tab:confusing_cls}
\end{table}

\noindent{\bf Implementations.}
We implement $\phi$ as a three-layer convolution network. We implement $\upsilon$ with a 2-layer fully connected network, which takes the concatenated instance and label as the input. All methods are optimized with Adam using an initial learning rate of 0.01 over 50 epochs. We set both the size of support and query set to 128.

\begin{table*}[t]
	\caption{Average few-shot classification accuracy on {\it Mini}ImageNet, {\it Tiered}ImageNet and CUB with ResNet-12 backbones over 10,000 trials. Methods with ResNet-18 backbone are marked with $\dagger$. }\label{tab:benchmark}
	\centering
	\begin{tabular}{@{\;}l@{\;}|@{\;}c@{\;}c@{\;}}
		\addlinespace
		\toprule
		{\it Mini}ImageNet  & 1-Shot 5-Way & 5-Shot 5-Way \\
		\midrule
		MAML{~\cite{FinnAL17Model}}                &58.37$\pm$0.49   & 69.76$\pm$0.46\\
		ProtoNet{~\cite{SnellSZ17Prototypical}}   &60.37$\pm$0.83   & 78.02$\pm$0.57  \\
		RelationNet{~\cite{Flood2017Learning}}    &61.40$\pm$0.83   & 77.00$\pm$0.64\\
		MetaOptNet{~\cite{Lee2019Meta}}          &62.64$\pm$0.61   & 78.63$\pm$0.46\\
		CAN{~\cite{Hou2019Cross}}             & 63.85$\pm$0.48   & 79.44$\pm$0.34  \\
		DeepEMD{~\cite{Zhang_2020_CVPR}}             & 65.91$\pm$0.82   & 82.41$\pm$0.56  \\
		FEAT{~\cite{ye2020fewshot}}             & 66.78$\pm$0.20  & 82.05$\pm$0.14  \\
		MeTAL{~\cite{baik2021meta}} & 59.64$\pm$0.38 & 76.20$\pm$0.19\\
		IMP+ZN{~\cite{fei2021z}}& 65.01$\pm$0.43 & 81.72$\pm$0.30\\
		CGCS{~\cite{gao2021curvature}}& 63.56$\pm$0.20 & 79.13$\pm$0.14\\
		MetaNAS{~\cite{wang2022global}}&64.24$\pm$0.11 & 79.75$\pm$0.13\\
		\midrule
		{\name} & \bf 67.32$\pm$0.20 & \bf 83.21$\pm$0.14 \\
		\bottomrule
	\end{tabular}
	\hfill
	\begin{tabular}{@{\;}l@{\;}|@{\;}c@{\;}c@{\;}}
		\addlinespace
		\toprule
		{\it Tiered}ImageNet & 1-Shot 5-Way & 5-Shot 5-Way \\
		\midrule
		MAML{~\cite{FinnAL17Model}}                &58.58$\pm$0.49   &71.27$\pm$0.43\\
		ProtoNet{~\cite{SnellSZ17Prototypical}}   &61.74$\pm$0.77   &80.00$\pm$0.55\\
		RelationNet{~\cite{Flood2017Learning}}    &54.48$\pm$0.93   & 71.32$\pm$0.78\\
		MetaOptNet{~\cite{Lee2019Meta}}           &65.99$\pm$0.72   & 81.56$\pm$0.63 \\
		AM3-TADAM{~\cite{Xing_2019_NIPS}}               & 69.08$\pm$0.47   & 82.58$\pm$0.31 \\
		DeepEMD{~\cite{Zhang_2020_CVPR}}             & 71.16$\pm$0.87  & 86.03$\pm$0.58  \\
		FEAT{~\cite{ye2020fewshot}}             & 70.80$\pm$0.23  & 84.79$\pm$0.16  \\
		MeTAL{~\cite{baik2021meta}} & 63.89$\pm$0.43 & 80.14$\pm$0.40\\
		IMP+ZN{~\cite{fei2021z}}& 67.58$\pm$0.51 & 83.94$\pm$0.35\\
		CGCS{~\cite{gao2021curvature}}& 71.66$\pm$0.23 & 85.50$\pm$0.15\\
		MetaNAS{~\cite{wang2022global}}&70.16$\pm$0.09 & 84.99$\pm$0.22\\
		\midrule
		{\name} & \bf 72.42$\pm$0.23 & \bf 86.60$\pm$0.16 \\
		\bottomrule
	\end{tabular}
	\hfill
	\begin{tabular}{@{\;}l@{\;}|@{\;}c@{\;}c@{\;}}
		\addlinespace
		\toprule
		CUB & 1-Shot 5-Way & 5-Shot 5-Way \\
		\midrule
		ProtoNet{~\cite{SnellSZ17Prototypical}}   &71.87$\pm$0.91   & 85.08$\pm$0.69 \\
		RelationNet{~\cite{Flood2017Learning}}    &62.45$\pm$0.98   & 76.11$\pm$0.69 \\
		SAML{~\cite{Hao_2019_ICCV}}               &69.33$\pm$0.22   & 81.56$\pm$0.15 \\
		DeepEMD{~\cite{Zhang_2020_CVPR}}         &75.65$\pm$0.83   & 88.69$\pm$0.50 \\
		Baseline++$^\dagger${~\cite{Chen2019Closer}}         &67.02$\pm$0.90   & 83.58$\pm$0.54\\
		AFHN$^\dagger${~\cite{li2020adversarial}}             & 70.53$\pm$1.01   & 83.95$\pm$0.63  \\
		Neg-Cosine$^\dagger${~\cite{liu2020negative}}         &72.66$\pm$0.85   & 89.40$\pm$0.43 \\
		Align$^\dagger${~\cite{afrasiyabi2020associative}}         &74.22$\pm$1.09   & 88.65$\pm$0.55\\
		IMP+ZN{~\cite{fei2021z}}& 71.22$\pm$0.48 & 85.51$\pm$0.31\\		CGCS-Ind{~\cite{gao2021curvature}}& 74.66$\pm$0.21 & 88.37$\pm$0.12\\
		CGCS-Trans{~\cite{gao2021curvature}}& 76.69$\pm$0.21 & 89.30$\pm$0.12\\
		\midrule
		{\name} & \bf 79.05$\pm$0.20 & \bf 90.85$\pm$0.11 \\
		\bottomrule
	\end{tabular}
\end{table*}

\noindent{\bf Results.}
Fig.~\ref{fig:semantics} (b) shows the t-SNE visualizations of two learned attribute maps by {\name}, where we can find the similarity between instances encoded by ``digit'' and ``color'', respectively. Results of classification accuracy per attribute are in Table~\ref{tab:confusing_cls}. Directly training a classifier over all classes leads to confusing results, where the color and digit patterns cannot not be differentiated. So SL leads to confusing results on two tasks. SL can be viewed as the ablation study for {\name}, which does not consider the possible influence of the multiple hidden attributes.
An ideal model would identify the latent attributes and make accurate predictions in both attributes.
CSL performs well for color but worse for digits. 
Since the maps discovered by {\name} capture the semantics of two latent attributes, {\name} achieves relatively high performance on each attribute by measuring the attribute-specific similarity.
The results validate the ability of {\name} to identify mixed latent attributes in a task and contextualize the model.

\begin{table}[t]
	\caption{Cross-domain few-shot classification performance evaluation over 10,000 trials of tasks. Both average mean accuracy and 95\% confidence interval are reported. Each model is trained over {\it Mini}ImageNet first and then evaluated over the novel classes from CUB. {\name} is implemented based on ResNet-12.}\label{tab:cross}
	\centering
	\begin{tabular}{l|cc}
		\addlinespace
		\toprule
		Setups $\rightarrow$ & 1-Shot 5-Way & 5-Shot 5-Way \\
		\midrule
		MAML{~\cite{FinnAL17Model}}                &34.01$\pm$1.25   & 48.83$\pm$0.62 \\
		ProtoNet{~\cite{SnellSZ17Prototypical}}   &33.27$\pm$1.09   & 52.16$\pm$0.17 \\
		MatchingNet{~\cite{VinyalsBLKW16Matching}}                &35.89$\pm$0.51   & 51.37$\pm$0.77 \\
		RelationNet{~\cite{Flood2017Learning}}    &42.44$\pm$0.77   & 57.77$\pm$0.69 \\
		MetaOptNet{~\cite{Lee2019Meta}}         &  44.79$\pm$0.75   & 64.98$\pm$0.68 \\
		
		\midrule
		{\name} &  \bf 44.82$\pm$0.87 & \bf 67.89$\pm$0.18 \\
		\bottomrule
	\end{tabular}
\end{table}

\subsection{Benchmark Few-Shot Classification}\label{sec:benchmark}
At last, we evaluate {\name} on few-shot classification benchmark, namely {\it Mini}ImageNet~\cite{VinyalsBLKW16Matching}, {\it Tiered}ImageNet~\cite{Ren2018Meta}, and Caltech-UCSD Birds-200-2011~(CUB)~\cite{WahCUB_200_2011}. 
We show that although there are no explicit attributes, we can improve the few-shot classification performance by manually setting the attributes, and the context-aware model improves the discerning ability.
We also evaluate the cross-domain few-shot classification ability to show whether a model could generalize to a novel domain.

\noindent{\bf Setups.}
{\it Mini}ImageNet contains 100 classes and 600 images per class. Following~\cite{ravi2016optimization,Ren2018Meta,ye2020fewshot}, there are 64 base classes, 16 classes for meta-validation, and 20 novel classes for meta-test. Similar to {\it Mini}ImageNet, {\it Tiered}ImageNet is a larger subset of ImageNet~\cite{RussakovskyDSKS15ImageNet} with a complicated hierarchical structure. It has 351/97/160 classes for meta-train/val/test~\cite{Ren2018Meta}. CUB contains fine-grained images for different species of birds, and there are 100/50/50 classes for meta-train/val/test~\cite{TriantafillouZU17Few}. 
All images are resized to 84 by 84 in advance.
The mean accuracy and 95\% confidence interval over 10,000 trials of tasks are computed. Each class in the query set contains 15 instances in the meta-training and meta-test stages.

\noindent{\bf Implementations.}
As mentioned in Section~\ref{sec:regularization}, we treat each base class as a latent attribute. Then the SAT variant of {\name} can infer the latent attribute when predicting each class by reformulating the $N$-way task to $N$ ``one-vs.-rest'' sub-tasks. 
In this case, each attribute corresponds to the property of a particular class. Hence, a particular few-shot support set is decomposed based on the properties of base classes first to construct the model with the contextualized embedding. Similarly, a novel class task is also decomposed into multiple components based on its similarity with base classes.
We set the number of latent attributes the same as the number of base classes in the meta-training set by default. 
The regularization $\Omega$ in Eq.~\ref{eq:stage1} is implemented as the cross-entropy between the base class attribute label and the output of the meta-disambiguation module $\nu$.
In particular, we can infer the latent attribute by the comparisons between instances from a particular class in the support set and instances from other classes in the support set. 
Thus, when we compute the similarity between novel instances, we measure them from multiple latent maps corresponding to the characteristics of base classes. The predictions of a particular class in the support set are based on those attribute-wise distances and are further weighted by the inferred latent attributes (\ie, the attribute number $C$ is the same as the base class size). 

ResNet-12~\cite{Lee2019Meta} is used to implement the encoder $\phi$ for both datasets. The backbone of the few-shot learner is pre-trained over the meta-training set of each dataset following~\cite{Rusu2018Meta,ye2020fewshot}. 
We use stochastic gradient descent with a momentum of 0.9 to optimize the few-shot learner. The initial learning rate is 0.0002, and the learning rate will time 0.1 after 100 tasks. 

\noindent{\bf Results.}
The mean accuracy and 95\% confidence interval over 10,000 trials are in Table~\ref{tab:benchmark}. Both 1-Shot 5-Way and 5-Shot 5-Way tasks are evaluated. We compare our method with recent ones like DeepEMD~\cite{Zhang_2020_CVPR} and FEAT~\cite{ye2020fewshot}. 
{\name} achieves promising results by identifying the mixed ``base class'' attributes during meta-training, and similarity between instances from novel classes would be decomposed over those attributes for better prediction during the meta-test.
It verifies contextualizing the meta-learning relates tasks and helps obtain a high-quality model.

We also investigate the robustness of the meta-learning model by testing the cross-domain results. In detail, a model trained on the meta-training set of the {\it Mini}ImageNet is evaluated on the meta-test set of CUB. Note that the best model is also selected based on the meta-validation set of {\it Mini}ImageNet. Such cross-domain evaluation is also named as out-of-distribution evaluation in~\cite{Lee2020Learning}. Therefore, a non-robust model is not able to generalize well across different domains. The results are listed in Table~\ref{tab:cross}. The performance of {\name} verifies its robustness with the contextualized meta-learning paradigm.

\begin{table}[t]
	\centering
	\caption{Ablation study over the few-shot classification performance on {\it Mini}ImageNet. Both average mean accuracy and 95\% confidence interval over 10,000 trials of tasks are reported. {\name} is implemented based on ResNet-12. We tune the balance weight of the regularizer $\Omega$ in Eq.~\ref{eq:stage1}.}
	\tabcolsep 2pt
	\begin{tabular}{c|ccccc}
		\addlinespace
		\toprule
		balance & 0     & 0.001 & 0.01  & 0.1   & 1 \\
		\midrule
		1-Shot & 66.84$\pm$0.20 & 67.32$\pm$0.20 & 67.13$\pm$0.20 & 66.95$\pm$0.20 & 54.57$\pm$0.20 \\
		5-Shot & 83.19$\pm$0.14 & 83.21$\pm$0.14 & 82.70$\pm$0.14 & 81.66$\pm$0.14 & 81.98$\pm$0.14 \\
		\bottomrule
	\end{tabular}
	\label{tab:ablation_balance}
\end{table}

\begin{table}[t]
	\centering
	\caption{Ablation study over the influence of meta disambiguation module of {\name} on {\it Mini}ImageNet. Both average mean accuracy and 95\% confidence interval over 10,000 trials of tasks are reported.}
	\begin{tabular}{c|cc|cc}
		\addlinespace
		\toprule
		 & \multicolumn{2}{c|}{{\it Mini}ImageNet} & \multicolumn{2}{c}{{\it Tiered}ImageNet}\\
		 & 1-Shot & 5-Shot & 1-Shot & 5-Shot \\
		\midrule
		{\name}$^-$ & 65.16$\pm$0.20 & 81.10$\pm$0.14 & 68.22$\pm$0.23 & 63.42$\pm$0.14\\
		{\name} & 67.32$\pm$0.20 & 83.21$\pm$0.14 & 72.42$\pm$0.23 & 86.60$\pm$0.14\\
		\bottomrule
	\end{tabular}
	\label{tab:ablation2}
\end{table}

\noindent{\bf Ablation studies.} 
We first show the influence of the regularization in Table~\ref{tab:ablation_balance}. The larger the balance weights, the more the regularizer matters. We have two observations. First, {\name} is able to discover the latent attributes corresponding to the base class without the regularization (the balance weight equals 0) and can get similar results with the best comparison methods (especially on 5-shot tasks). Second, by adding the regularization, {\name} models the attribute better with the introduced supervision, and the few-shot classification performance can be further improved.

Furthermore, we compare {\name} with the variant {\name}$^-$ without the meta disambiguation module on {\it Mini}ImageNet and {\it Tiered}ImageNet. We find the performance of {\name}$^-$ drops a lot w.r.t. {\name}. The results verify that weighting the influence of the learned hidden attributes is essential for few-shot learning.

\begin{figure}[t]
	\centering
	\includegraphics[width=0.9\linewidth]{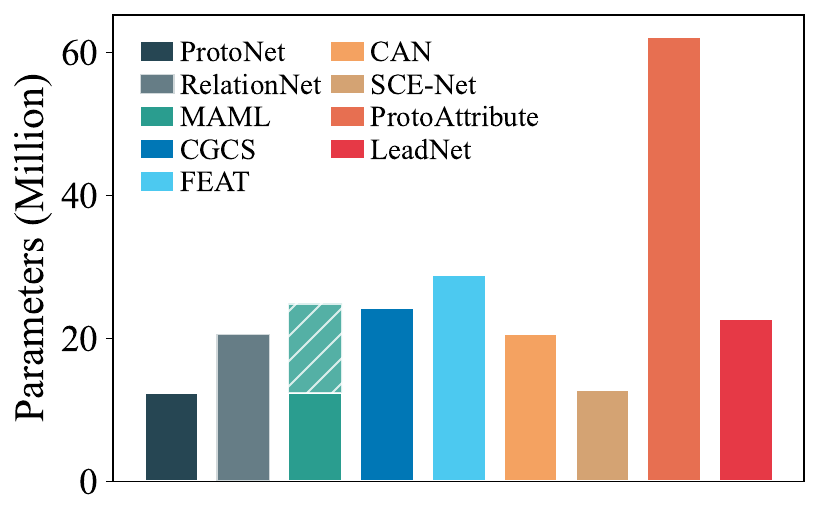}
	\caption{Model size comparison of different methods. The shaded area in MAML represents the parameters used during training but dropped during inference.} \label{fig:param_num}
\end{figure}

\noindent{\bf Model Size Comparisons.} Compared to a vanilla backbone, {\name} requires extra parameters in three aspects to learn the attribute-specific embeddings, \ie, attribute-specific linear mappings $\mathcal{L}_C=\{L_1, \ldots, L_C\}$,  local map identifier $h(\cdot)$, and context attribute identifier $\upsilon(\mathcal{\cdot})$. Among them, each mapping $L_c$ is a simple linear layer, while $h(\cdot)$ and $\upsilon(\mathcal{\cdot})$ are implemented with two-layer MLPs. To quantitatively show the model size, we report the number of parameters of different methods in Figure~\ref{fig:param_num}. 
Among them, MAML requires saving an extra gradient matrix during the meta-training process, which is the same scale as the backbone size. We show those parameters with shaded areas. 
ProtoAttribute trains a separate backbone for each attribute, with the largest model size among all methods. Other methods like RelationNet, FEAT, and CAN adopt extra modules, \eg, similarity measure network or self-attention blocks, which consume additional parameters. 
As we can infer from the figure, {\name} shares the same parameter scale as other compared methods while achieving the best performance.

\section{Conclusion}
{\name} is a {\em unifying} approach that aims to reduce semantic ambiguity in meta-learning when instances in a support set possess one or more latent attributes.
By leveraging the change in similarity relationships between instances across different contexts, {\name} effectively uncovers the underlying latent attributes associated with the instances. 
The meta-learned {\em context-aware} embeddings enhance the discerning ability and improve the robustness of the meta-model against out-of-distribution examples as well as (pseudo) tasks.
Experiments verify {\name}'s ability to decompose meaningful latent embedding spaces for a variety of ``downstream'' applications.
Future work includes applying {\name} across modalities and domains.

\ifCLASSOPTIONcompsoc
  \section*{Acknowledgments}
\else
  \section*{Acknowledgment}
\fi

This work is partially supported by the National Key R\&D Program of China (2022ZD0114805), NSFC (62376118, 62176117, 62272231), NSF of Jiangsu Province
(BK20200313, BK20210340), Collaborative Innovation Center of Novel Software Technology and Industrialization, the Fundamental Research Funds for the Central Universities (No. NJ2022028).

\ifCLASSOPTIONcaptionsoff
  \newpage
\fi





{\small
\bibliographystyle{IEEEtran}
\bibliography{paper}}

\newpage
\section{Supplementary Experiments}
\label{sec:supp_experiment}

\subsection{Regression with Random Selected Component} \label{sec:regress2}
The few-shot regression task requires a model to fit a curve based on limited given continuous labeled examples in a support set. We investigate a more difficult scenario with multiple curve families, and a few-shot regression task could be sampled from one of those latent attributes. This is an {\bf SAT} setting, which requires the model to adapt quickly from pseudo tasks without the supervision of attribute labels. 

\noindent\textbf{Setups}. We follow~\cite{Yao2019Hierarchically} and sample curves from one of the four function families, \ie, (a) Sinusoids: $y(x) = A\sin(wx) + b$, $A\sim U[0.1, 5.0]$, $w\sim U[0.8, 1.2]$, and $b\sim U[0, 2\pi]$; (b) Linear: $y(x)=Ax+b$, $A\sim U[-3.0, 3.0]$, and $b\sim U[-3.0, 3.0]$; (c) Quadratic: $y(x)=Ax^2+bx+c$, $A\sim U[-0.2, 0.2]$, $b\sim U[-2.0, 2.0]$, and $c\sim U[-3.0, 3.0]$. (d) Cubic: $y(x)=Ax^3+bx^2+cx+d$, $A\sim U[-0.1, 0.1]$, $b\sim U[-0.2, 0.2]$, $c\sim U[-2.0, 2.0]$, and $d\sim U[-3.0, 3.0]$. $U[\cdot, \cdot]$ denotes a uniform sampler. In other words, we first randomly select one function family, and then randomly sample corresponding parameters from the uniform distribution. 5 or 10 examples are sampled from the curve as the support set, and non-overlapping 100 instances act as the query set. We use the Mean Square Error (MSE) to measure the quality of the predictions on the query set (the lower the better).

\begin{table}[t]
	\caption{Mean Square Error (MSE) on the few-shot regression tasks with randomly selected component over 4,000 tasks.}\label{tab:regression1}
	\centering
	\begin{tabular}{l|cc}
		\addlinespace
		\toprule
		Method  & 5-Shot  & 10-Shot \\
		\midrule
		MAML{~\cite{FinnAL17Model}}                &2.205$\pm$0.121   & 0.761$\pm$0.068\\
		ProtoNet{~\cite{SnellSZ17Prototypical}}                &1.904$\pm$0.080   & 0.876$\pm$0.083\\
		Meta-SGD{~\cite{LiZCL17Meta}}                &2.053$\pm$0.117   & 0.836$\pm$0.065\\
		MT-Net{~\cite{Lee2018Gradient}}  & 2.435$\pm$0.130 & 0.967$\pm$0.056 \\
		BMAML{~\cite{Yoon2018Bayesian}}  & 2.016$\pm$0.109 & 0.698$\pm$0.054 \\
		MuMoMAML{~\cite{Vuorio2019Multimodal}}  & 1.096$\pm$0.085  & 0.256$\pm$0.028 \\
		HSML{~\cite{Yao2019Hierarchically}}  & 0.856$\pm$0.073  & 0.161$\pm$0.021 \\
		\midrule
		{\name}  & \bf 0.776$\pm$0.088 & \bf 0.121$\pm$0.023 \\
		\bottomrule
	\end{tabular}
\end{table}

\begin{figure*}[t]
	\begin{center}
		\subfigure[Ground-Truth, Attribute-{1}]
		{\includegraphics[width=.24\textwidth]{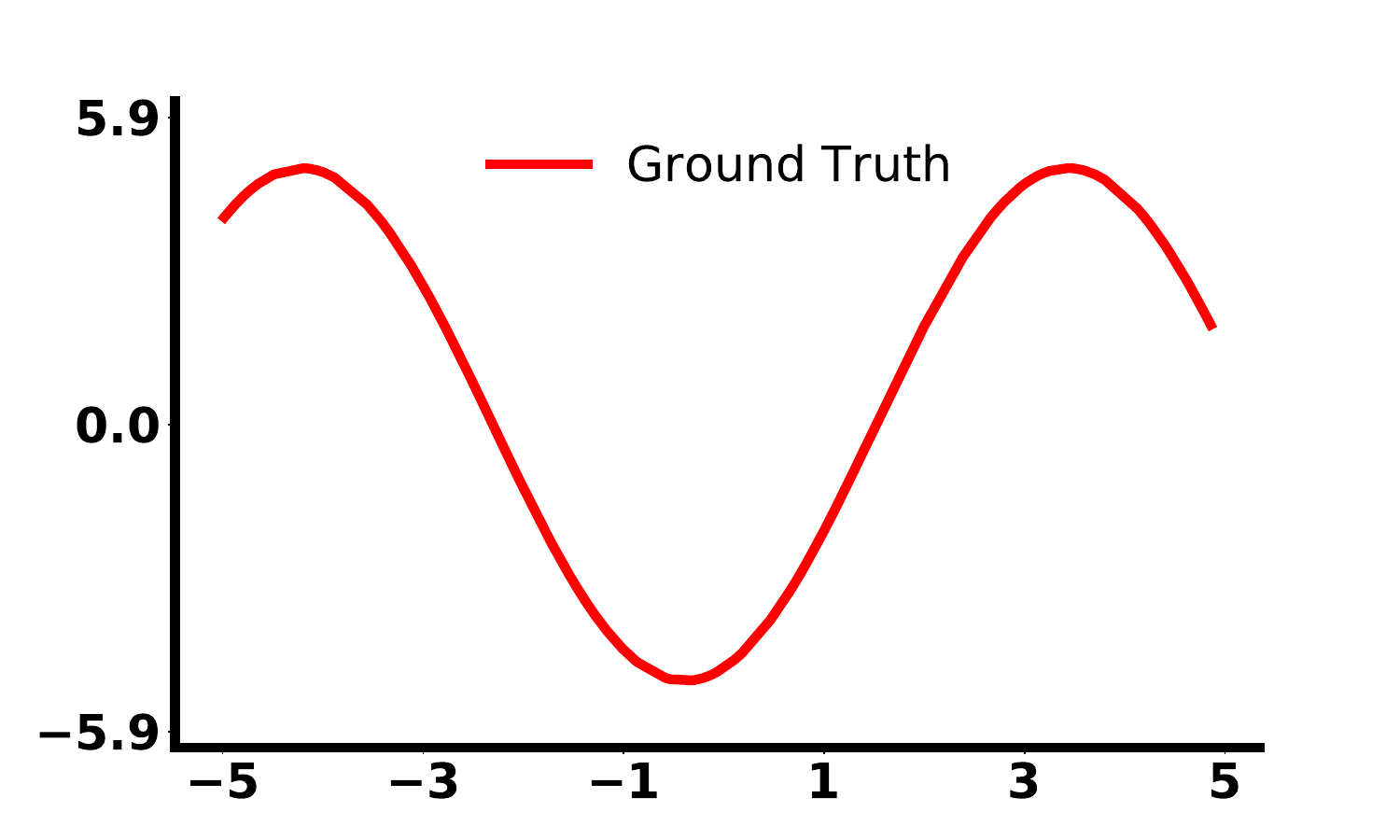}}
		\subfigure[Ground-Truth, Attribute-{2}]
		{\includegraphics[width=.24\textwidth]{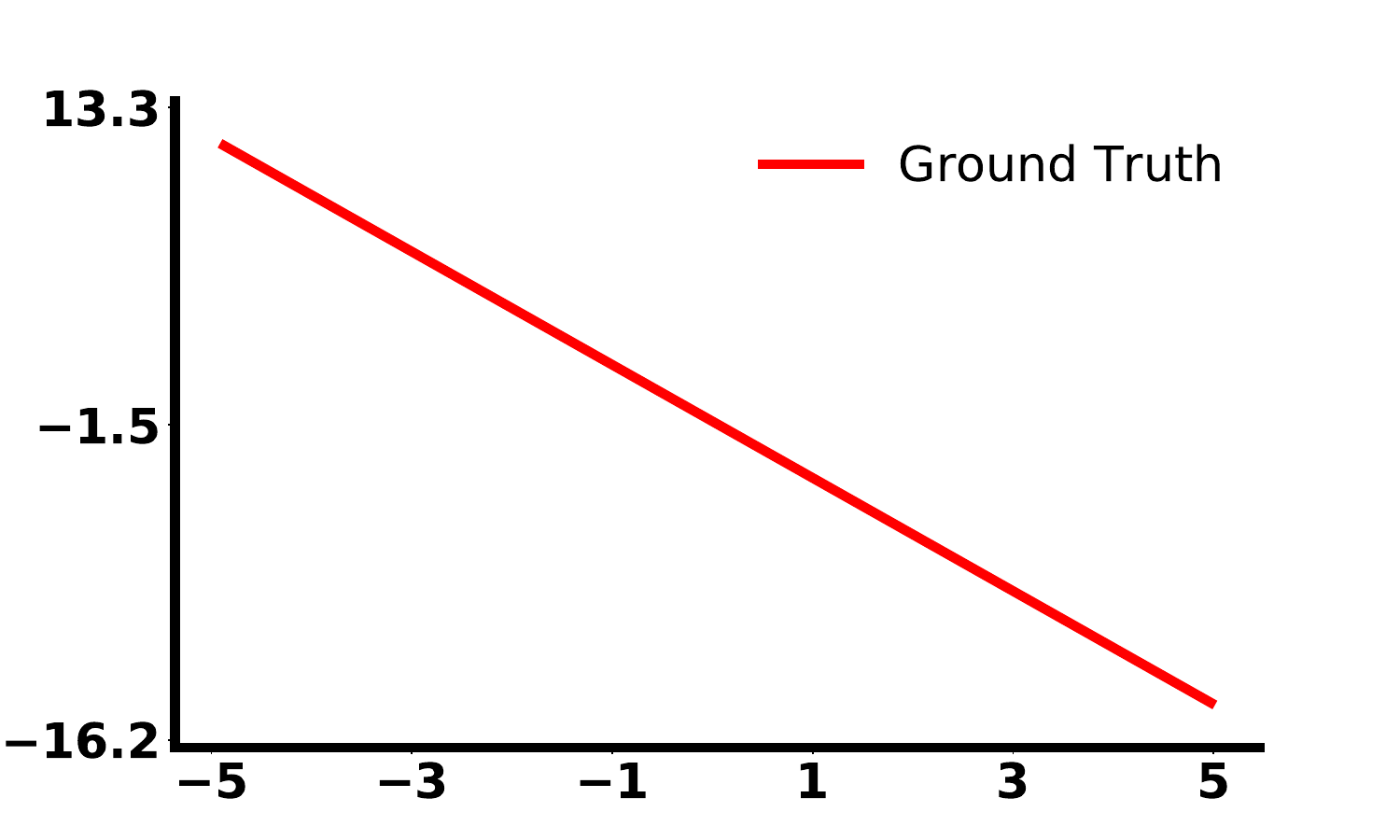}}
		\subfigure[Ground-Truth, Attribute-{3}]
		{\includegraphics[width=.24\textwidth]{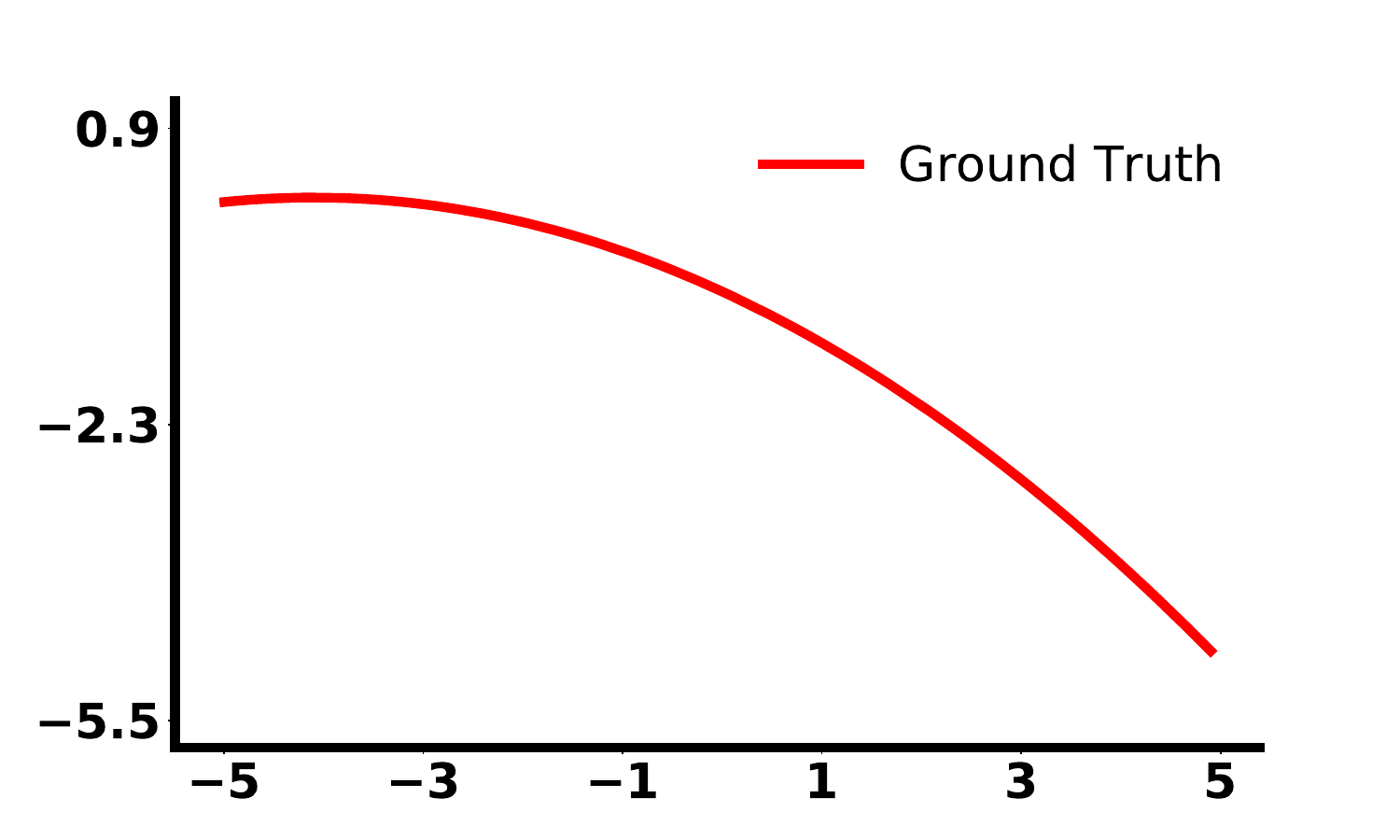}}
		\subfigure[Ground-Truth, Attribute-{4}]
		{\includegraphics[width=.24\textwidth]{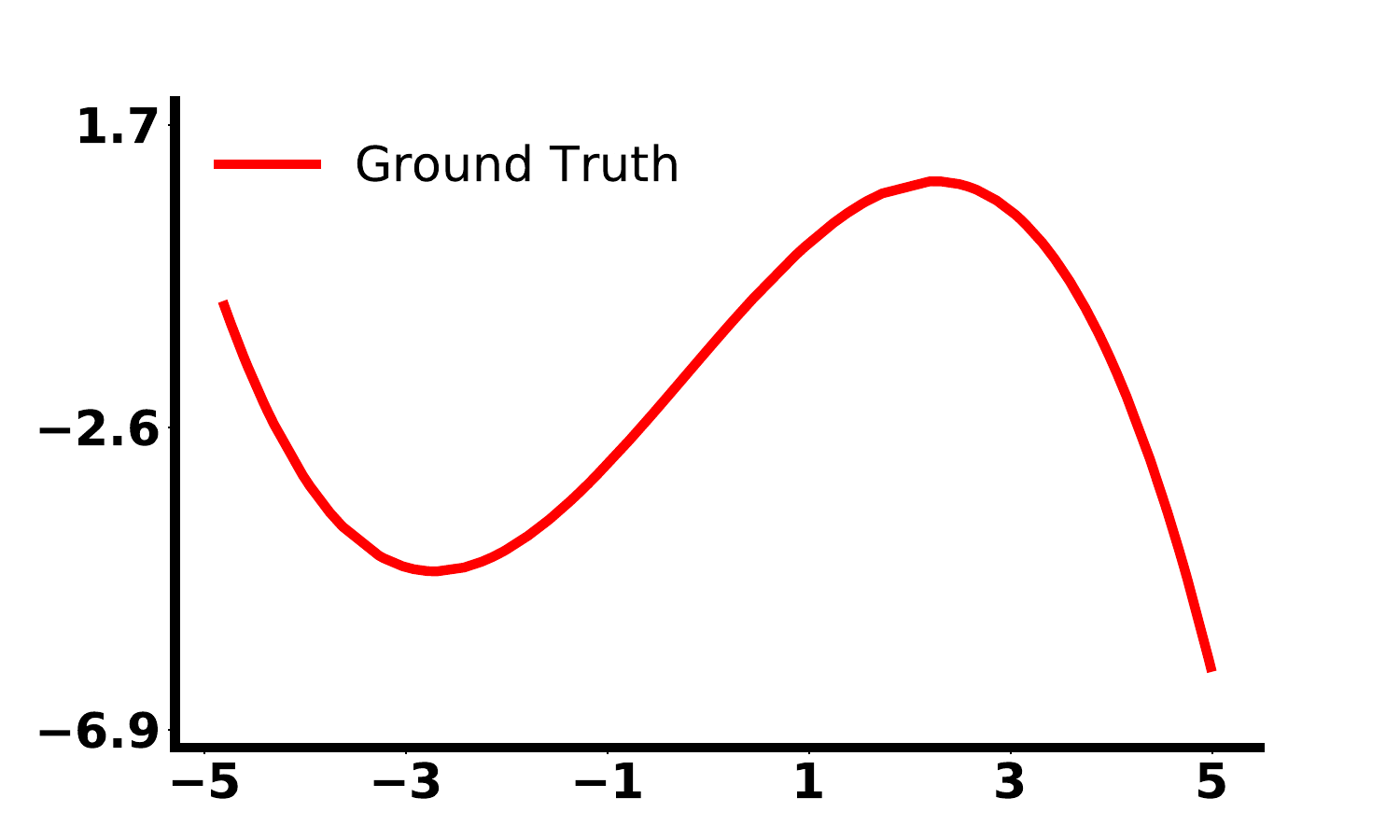}}
		\\
		\subfigure[\name, Attribute-{1}]
		{\includegraphics[width=.24\textwidth]{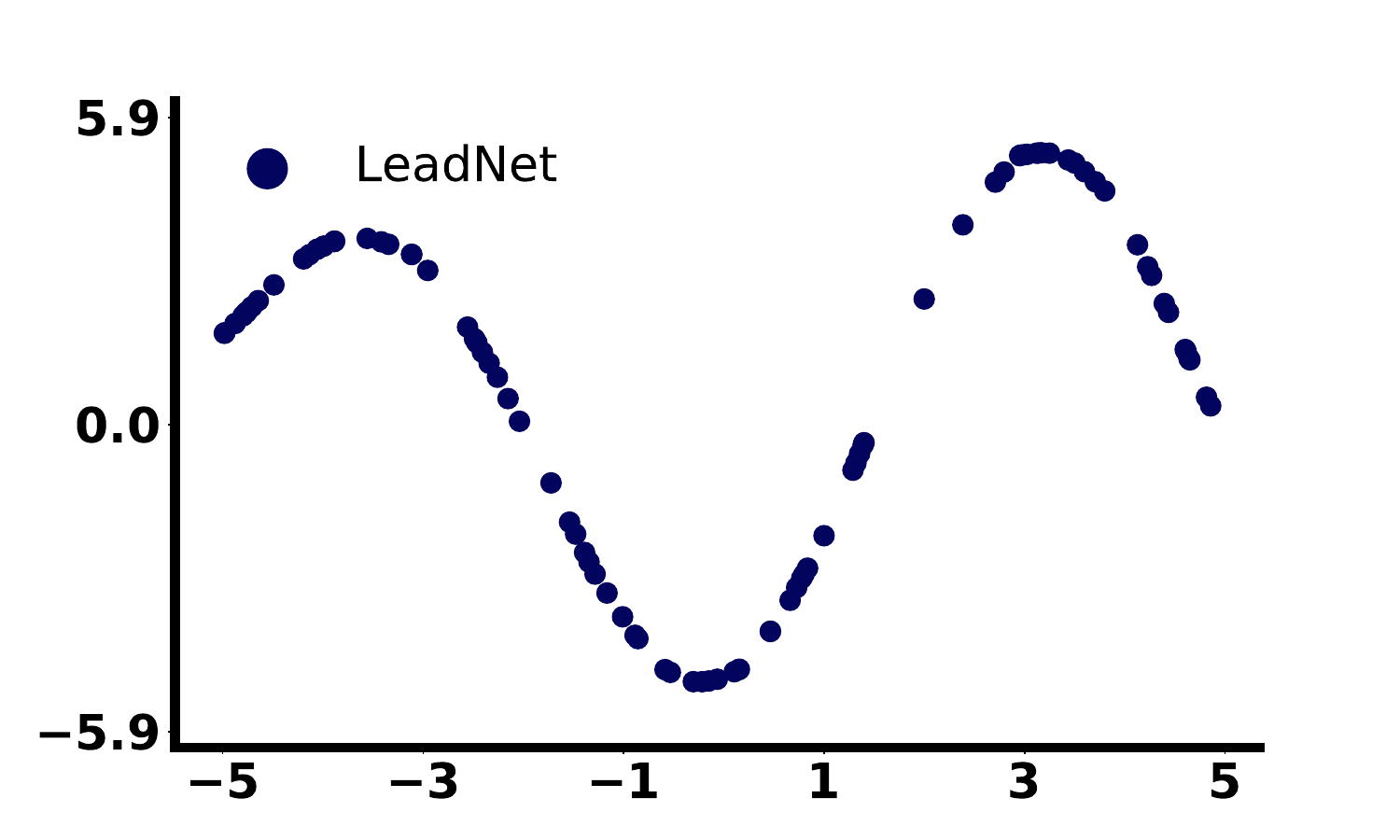}}
		\subfigure[{\name}, Attribute-{2}]
		{\includegraphics[width=.24\textwidth]{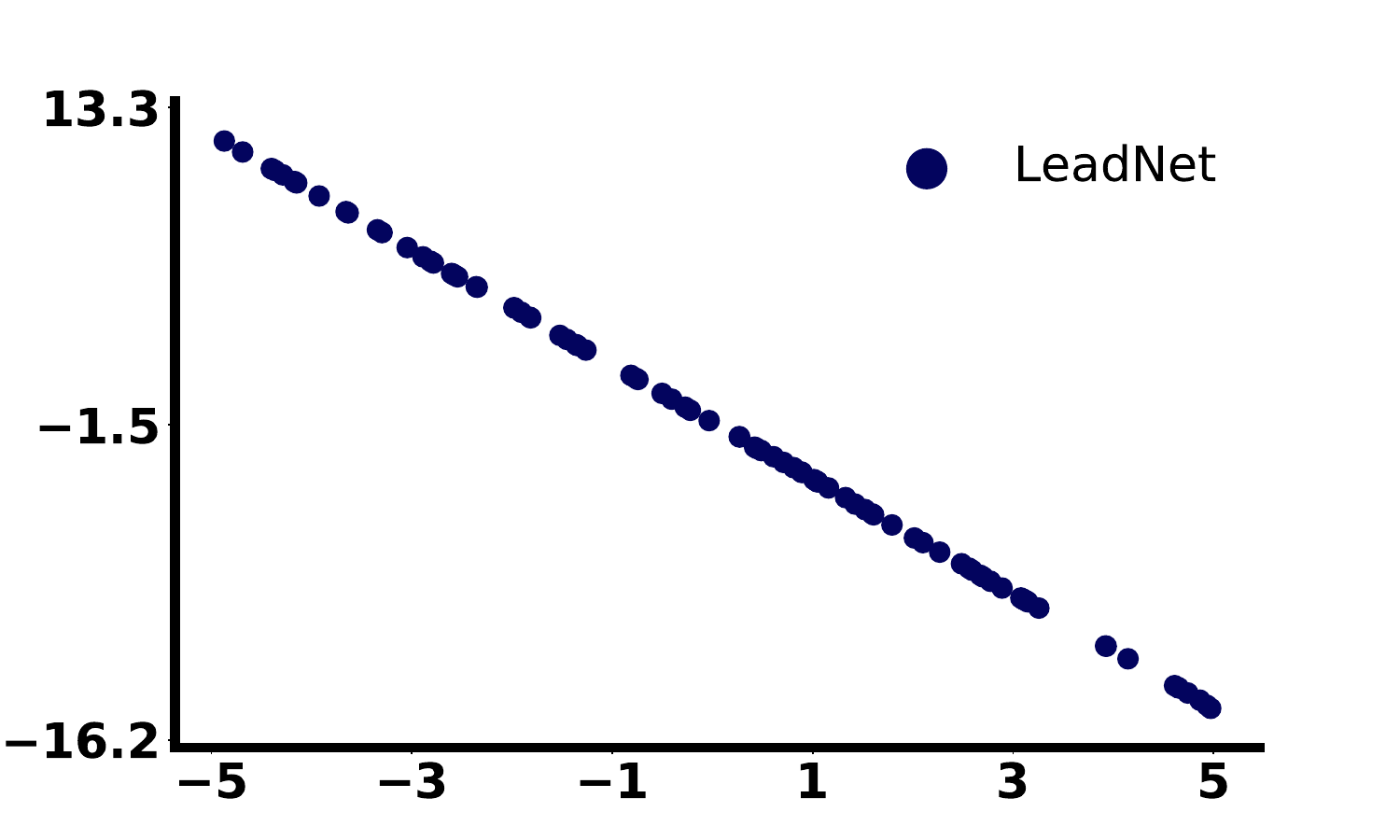}}
		\subfigure[{\name}, Attribute-{3}]
		{\includegraphics[width=.24\textwidth]{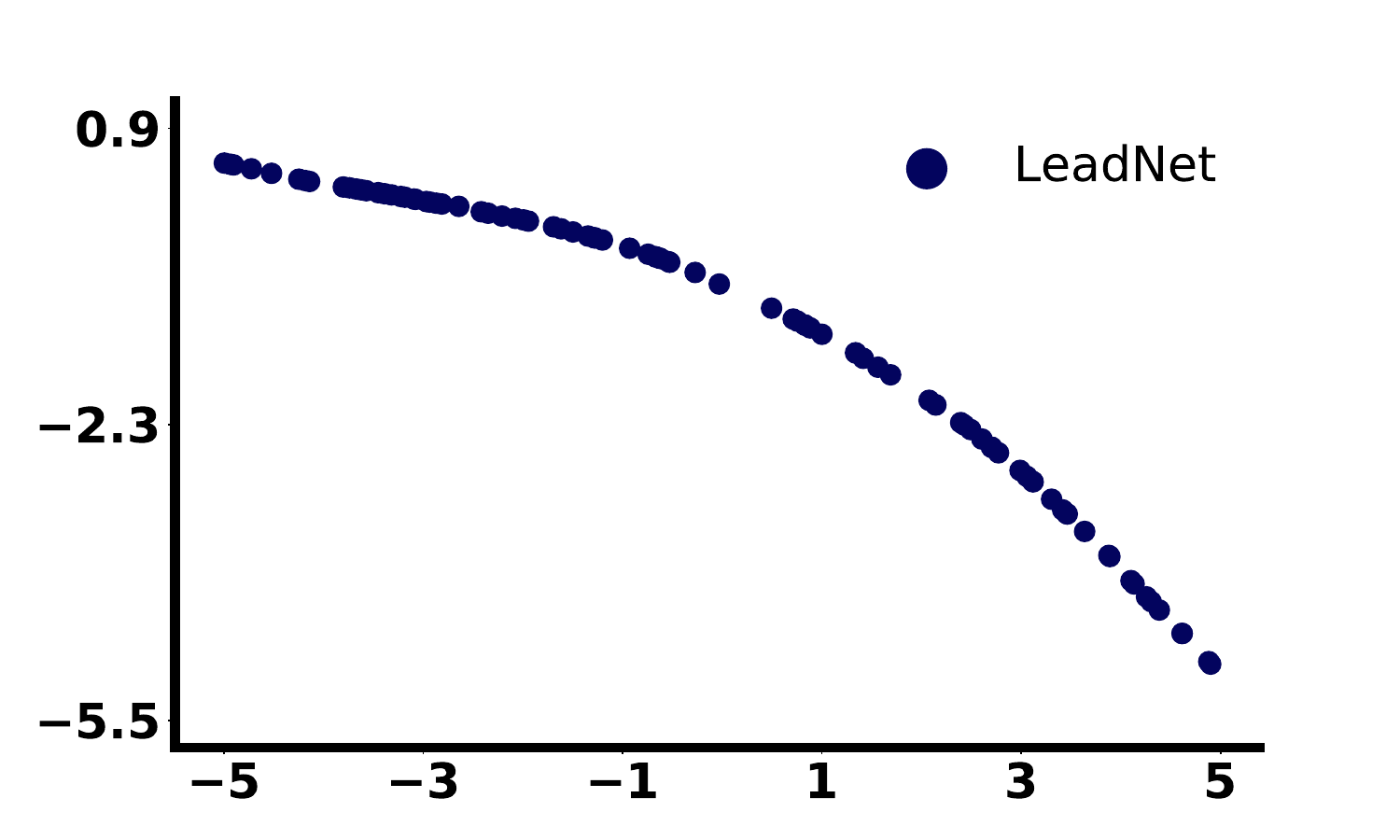}}
		\subfigure[{\name}, Attribute-{4}]
		{\includegraphics[width=.24\textwidth]{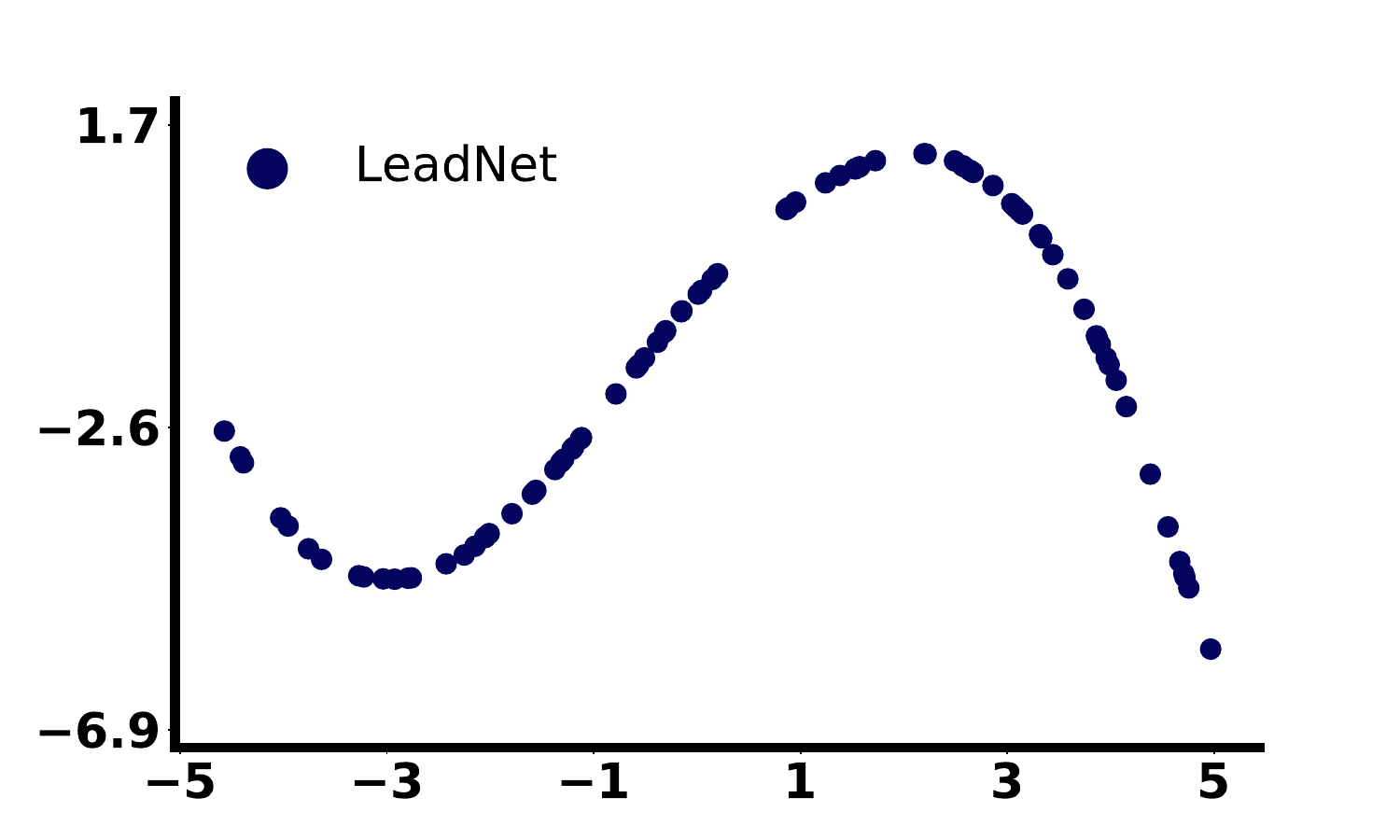}}\\
		\subfigure[ProtoNet, Attribute-{1}]
		{\includegraphics[width=.24\textwidth]{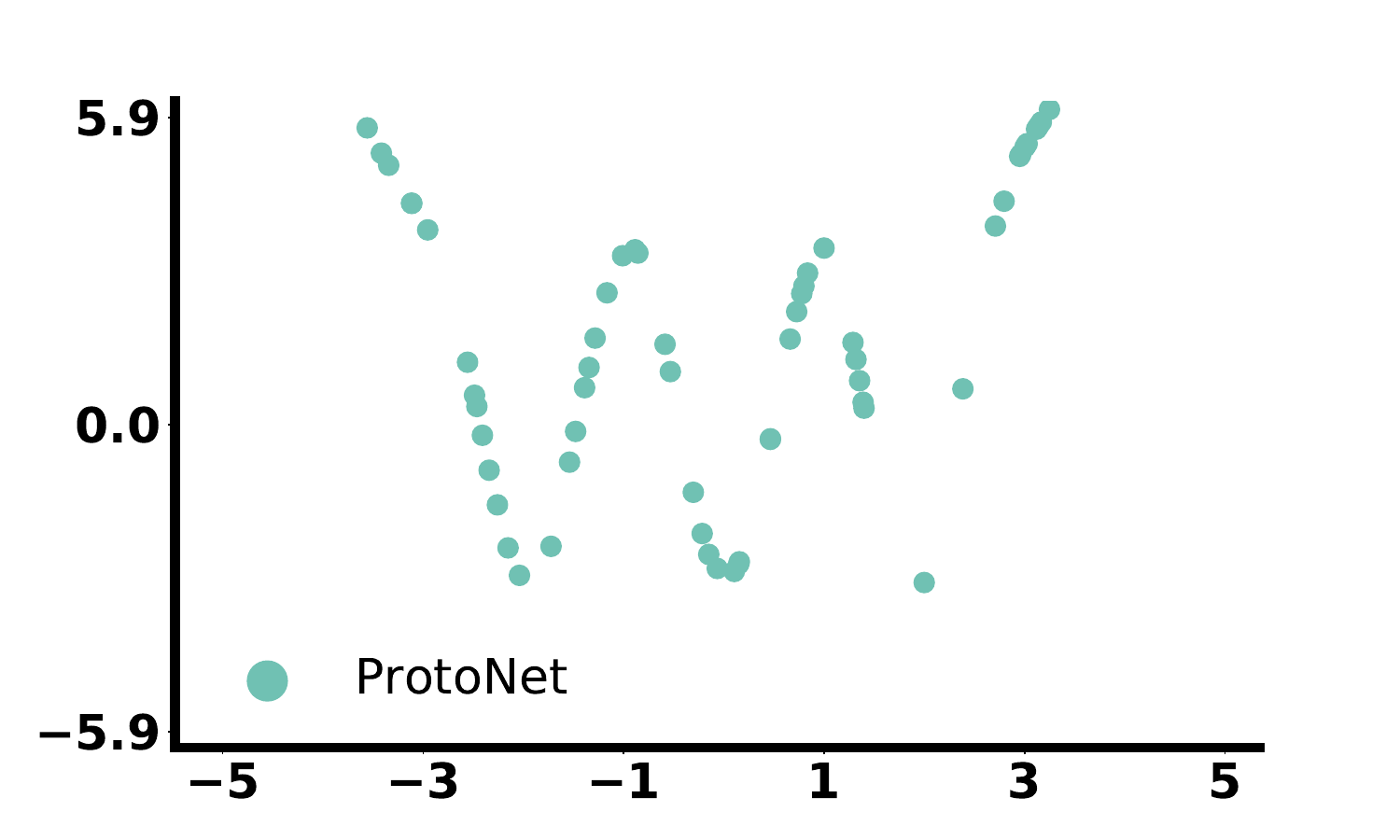}}
		\subfigure[ProtoNet, Attribute-{2}]
		{\includegraphics[width=.24\textwidth]{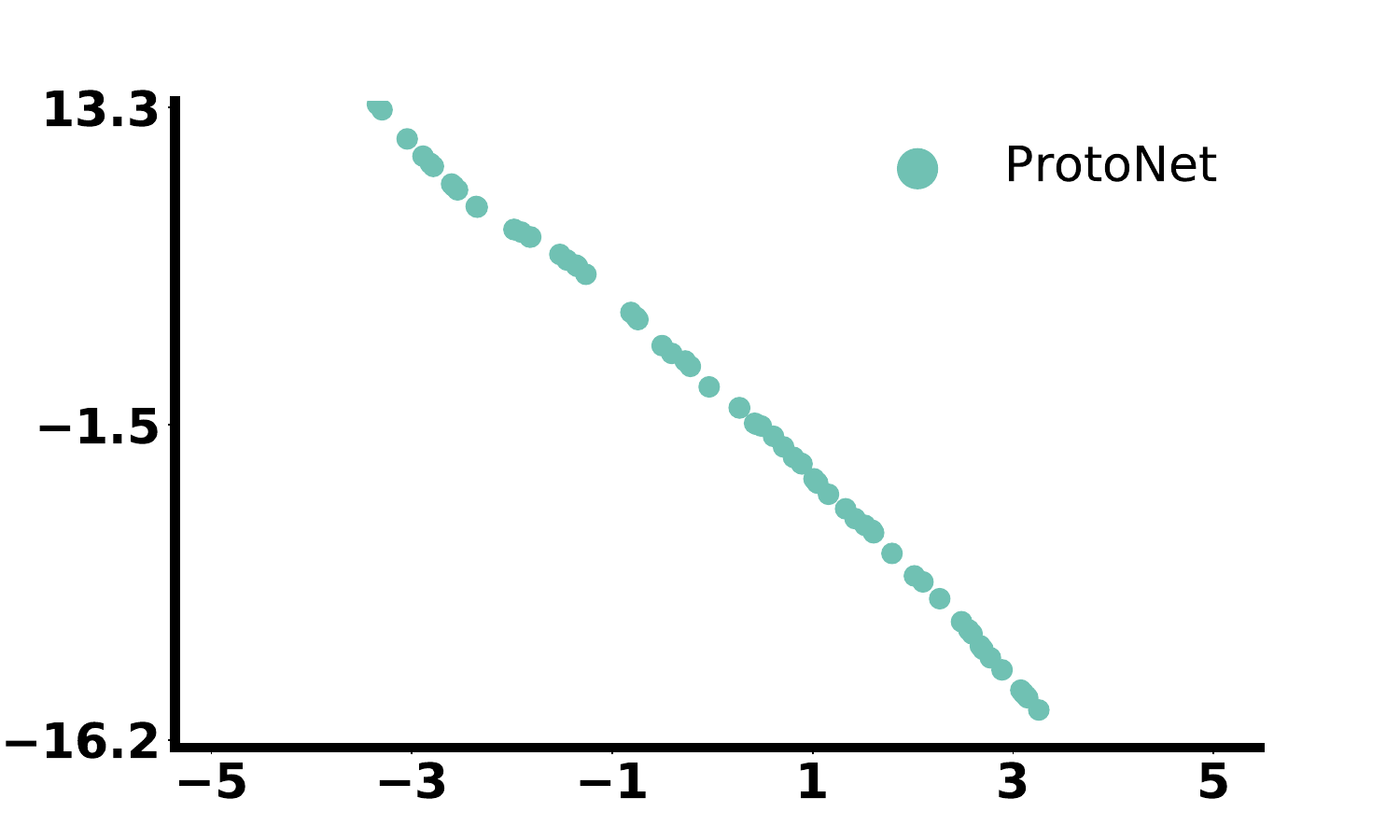}}
		\subfigure[ProtoNet, Attribute-{3}]
		{\includegraphics[width=.24\textwidth]{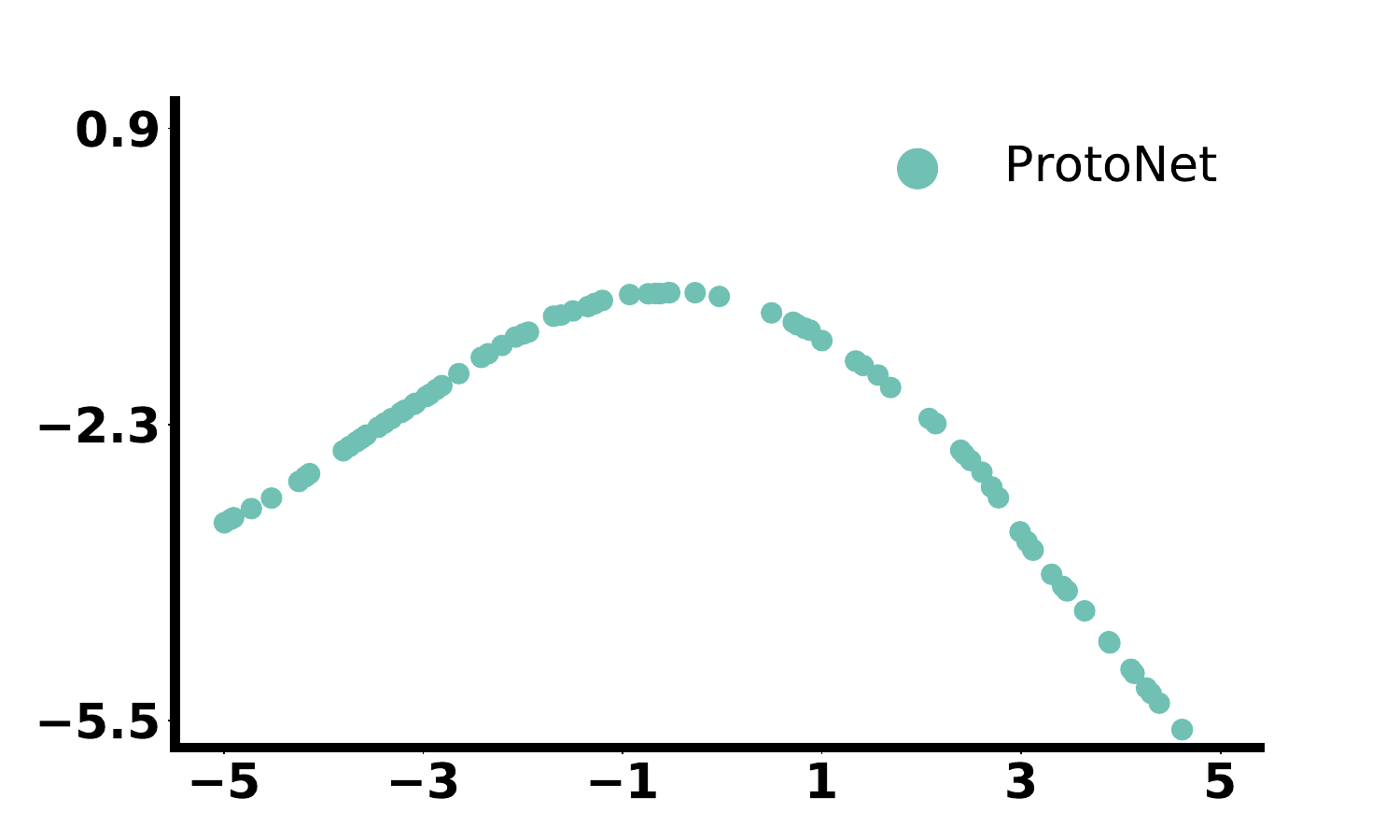}}
		\subfigure[ProtoNet, Attribute-{4}]
		{\includegraphics[width=.24\textwidth]{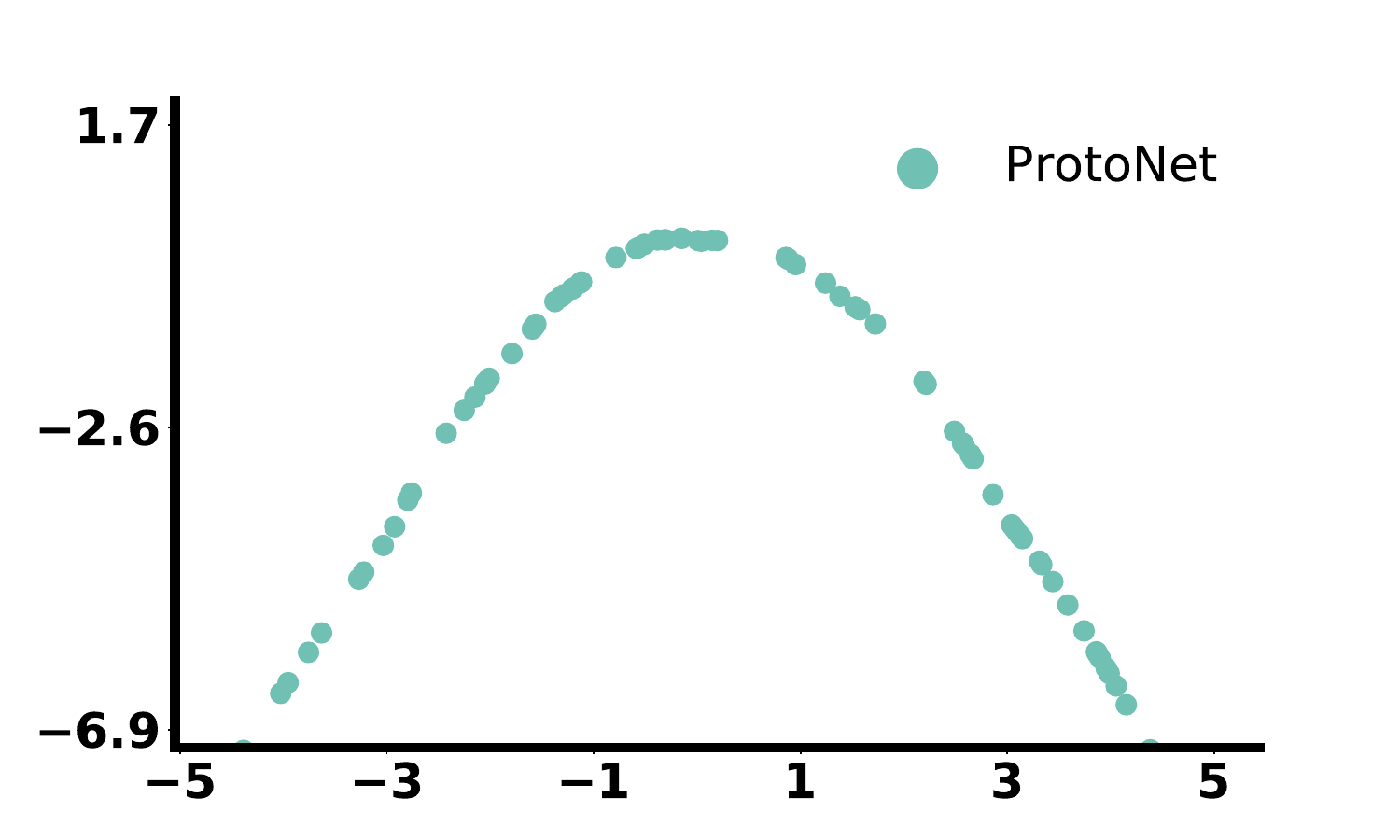}}
	\end{center}
	\caption{Visualization of a few-shot regression task with four different curve families (column 1-4, sinusoids, linear, quadratic, and cubic). A few-shot task is sampled from one of the latent attributes, and a model is required to fit the curve based on 5-shot training examples.
		The top row, second row and the bottom row correspond to the ground-truth and the fitted curves of {\name} and ProtoNet, respectively. {\name} is able to contextualize the model and discover all latent attributes successfully.
	} \label{fig:regression2}
\end{figure*}

\noindent\textbf{Implementations}. Similar to Section~5.1, we use the square loss in the objective. We implement $\phi$ as a three-layer fully connected network. Since labels are real-valued, we treat all instances in a task as a tuple and generate its task-specific probability based on the fully connected networks. We use the same optimization hyper-parameters with~\cite{Yao2019Hierarchically}.

\noindent\textbf{Results}.
We compare {\name} with various task-adaptive few-shot learning methods like MuMoMAML~\cite{Vuorio2019Multimodal} and HSML~\cite{Yao2019Hierarchically}, which takes the latent components into account during meta-learning. The MSE over 4000 trials are reported in Table~\ref{tab:regression1}. We find the task-adaptive methods can get better performance than others. Our {\name} achieves lower MSE values in both 5-shot and 10-shot cases. 
We also visualize the regression results of {\name} in Fig.~\ref{fig:regression2}, which contains the ground-truth components (upper) and the fitted curve by {\name} (middle). 
{\name} infers the latent attribute of a few-shot task and then predicts the values for the query set. For comparison, we also visualize the prediction of the baseline method, \ie, ProtoNet, at the bottom. ProtoNet does not consider the multiple attributes and only assigns the prediction w.r.t. the similarity of instance pairs, which gets inferior performance. For comparison, we find the predicted curves by {\name} are similar to the ground-truth.
The superiority of {\name} verifies its ability to identify the right attribute of a regression task and contextualize the model accordingly.

\begin{table}[t]
	\caption{Cross-domain few-shot classification performance evaluation over 10,000 trials of tasks. Both average mean accuracy and 95\% confidence interval are reported. Each model is trained over {\it Mini}ImageNet first and then evaluated over the novel classes from CUB. All methods are implemented based on ResNet-10 with $224*224$ inputs.}\label{tab:cross_res10}
	\centering
	\begin{tabular}{l|cc}
		\addlinespace
		\toprule
		Setups $\rightarrow$ & 1-Shot 5-Way & 5-Shot 5-Way \\
		\midrule
		FWT{~\cite{tseng2020cross}}                &47.47$\pm$0.75   &66.98$\pm$0.68 \\
		ATA{~\cite{wang2021cross}}   &50.26$\pm$0.50   & 65.31$\pm$0.40 \\
		wave-SAN{~\cite{fu2022wave}}                &50.33$\pm$0.73   & \bf 71.16$\pm$0.66 \\
		\midrule
		{\name} &  \bf 50.59$\pm$0.61 &  70.88$\pm$0.32 \\
		\bottomrule
	\end{tabular}
\end{table}

\subsection{Cross-Domain Few-Shot Learning}
In the main paper, we conduct experiments for cross-domain few-shot learning with ResNet12. Apart from ResNet12, there is another commonly adopted backbone, \ie, ResNet10. Hence, we supply the results with ResNet10 in this section. Specifically, we compare to FWT~\cite{tseng2020cross}, ATA~\cite{wang2021cross} and wave-SAN~\cite{fu2022wave} in this setting, and report the results in Table~\ref{tab:cross_res10}. As shown in the Table, {\name} performs competitively if fairly compared to these works. It achieves the best performance in 1-shot scenario and runner-up in 5-shot scenario. Although {\name} is not specially designed for the cross-domain few-shot learning task, it still shows strong capability in transferring to downstream tasks.

%

%

\end{document}